\documentclass[10pt,twocolumn,letterpaper]{article}
\usepackage[pagenumbers]{cvpr}

\usepackage{adjustbox}

\usepackage{dblfloatfix}
\usepackage{enumitem}
\setlist[itemize]{leftmargin=*, noitemsep, topsep=2pt}
\setlist[enumerate]{leftmargin=*, noitemsep, topsep=2pt}
\usepackage{placeins}            

\newcommand{\bk}[2]{#1~\cite{#2}}

\newcommand{\tabtwostyle}{%
  \scriptsize
  \setlength{\tabcolsep}{3.2pt}%
\renewcommand{\arraystretch}{1.15}%
}

\newenvironment{TabTwo}[1][t]{%
  \begin{table*}[#1]
  \centering
  \begingroup
  \tabtwostyle
}{%
  \endgroup
  \end{table*}
}

\newcommand{\best}[1]{\textbf{#1}}
\usepackage{graphicx}
\usepackage[T1]{fontenc}
\usepackage[utf8]{inputenc}
\usepackage{amssymb}

\usepackage{algorithm}
\usepackage{lipsum}

\usepackage{mathtools}

\usepackage{multirow}

\usepackage{booktabs}
\usepackage{tabularx}
\usepackage{array}
\usepackage{xurl}        

\newcolumntype{L}{>{\raggedright\arraybackslash}X}
\newcolumntype{C}[1]{>{\centering\arraybackslash}p{#1}}

\newcommand{\dataset}[1]{\textsc{#1}}
\newcommand{\dslink}[2]{\href{\detokenize{#2}}{\dataset{#1}}}

\usepackage{listings}
\usepackage{upquote}  
\usepackage{xcolor,colortbl}
\lstdefinestyle{overleaf}{
    backgroundcolor=\color[rgb]{0.95,0.95,0.92},   
    commentstyle=\color[rgb]{0,0.6,0},
    keywordstyle=\color{magenta},
    numberstyle=\tiny\color[rgb]{0.5,0.5,0.5},
    stringstyle=\color[rgb]{0.58,0,0.82},
    basicstyle=\ttfamily\footnotesize,
    breakatwhitespace=false,         
    breaklines=true,                 
    captionpos=b,                    
    keepspaces=true,                 
    numbers=left,                    
    numbersep=5pt,                  
    showspaces=false,                
    showstringspaces=false,
    showtabs=false,                  
    tabsize=2
}
\lstdefinestyle{mocov3}{
  backgroundcolor=\color{white},
  basicstyle=\fontsize{7.5pt}{7.5pt}\ttfamily\selectfont,
  columns=fullflexible,
  breaklines=true,
  captionpos=b,
  commentstyle=\fontsize{7.5pt}{7.5pt}\color[rgb]{0.25,0.5,0.5},
  keywordstyle=\fontsize{7.5pt}{7.5pt}\color[rgb]{0.85,0.18,0.50},
}
\usepackage{etoolbox}
\makeatletter
\AfterEndEnvironment{algorithm}{\let\@algcomment\relax}
\AtEndEnvironment{algorithm}{\kern2pt\hrule\relax\vskip3pt\@algcomment}
\let\@algcomment\relax
\newcommand\algcomment[1]{\def\@algcomment{\footnotesize#1}}
\renewcommand\fs@ruled{\def\@fs@cfont{\bfseries}\let\@fs@capt\floatc@ruled
  \def\@fs@pre{\hrule height.8pt depth0pt \kern2pt}%
  \def\@fs@post{}%
  \def\@fs@mid{\kern2pt\hrule\kern2pt}%
  \let\@fs@iftopcapt\iftrue}
\makeatother

\usepackage[pagebackref,breaklinks,colorlinks]{hyperref}

\usepackage[capitalize]{cleveref}
\crefname{section}{Sec.}{Secs.}
\Crefname{section}{Section}{Sections}
\Crefname{table}{Table}{Tables}
\crefname{table}{Tab.}{Tabs.}

\newcommand{\model}[1]{\textbf{#1}}

\def\cvprPaperID{0000} 
\def\confName{CVPR}
\def\confYear{2023}

\newcommand{\authsep}{\;\;}

\begin{document}
\raggedbottom

\title{DinoDental: Benchmarking DINOv3 as a Unified Vision Encoder for Dental Image Analysis}

\author{
Kun Tang$_{1}$ \authsep
Xinquan Yang$_{1}$ \authsep
Mianjie Zheng$_{1}$ \authsep
Xuefen Liu$_{1}$ \authsep
Xuguang Li$_{2}$ \authsep
\\
Xiaoqi Guo$_{1}$ \authsep
Ruihan Chen$_{3}$ \authsep
Linlin Shen$_{1}$ \authsep
He Meng$_{2}$ 
}
\makeatletter
\maketitle
{\let\thefootnote\relax\footnote{%
\hspace*{-1.8em}\textsuperscript{1} Shenzhen University.\\
\textsuperscript{2} Shenzhen University General Hospital.\\
\textsuperscript{3} Chongqing University.
}}
\makeatother


\vspace{-1em}
\begin{abstract}

The scarcity and high cost of expert annotations in dental imaging present a significant challenge for the development of AI in dentistry. DINOv3, a state-of-the-art, self-supervised vision foundation model pre-trained on 1.7 billion images, offers a promising pathway to mitigate this issue. However, its reliability when transferred to the dental domain—with its unique imaging characteristics and clinical subtleties—remains unclear.
To address this, we introduce DinoDental, a unified benchmark designed to systematically evaluate whether DINOv3 can serve as a reliable, off-the-shelf encoder for comprehensive dental image analysis without requiring domain-specific pre-training. Constructed from multiple public datasets, DinoDental covers a wide range of tasks, including classification, detection, and instance segmentation on both panoramic radiographs and intraoral photographs. We further analyze the model's transfer performance by scaling its size and input resolution, and by comparing different adaptation strategies, including frozen features, full fine-tuning, and the parameter-efficient  Low-Rank Adaptation (LoRA) method. Our experiments show that DINOv3 can serve as a strong unified encoder for dental image analysis across both panoramic radiographs and intraoral photographs, remaining competitive across tasks while showing particularly clear advantages for intraoral image understanding and boundary-sensitive dense prediction. 
Collectively, DinoDental provides a systematic framework for comprehensively evaluating DINOv3 in dental analysis, establishing a foundational benchmark to guide efficient and effective model selection and adaptation for the dental AI community.
\end{abstract}

\setlength{\parindent}{0pt}
\setlength{\parskip}{2pt} 

\section{Introduction}\label{sec:intro}

Dental image analysis represents a promising yet challenging frontier for artificial intelligence (AI) in medicine~\cite{huang2023dl_dentistry,schwendicke2020ai}. The clinical demand is substantial, with oral diseases affecting billions globally and untreated dental caries remaining among the most prevalent health conditions worldwide~\cite{who_oral_health_factsheet,who_oral_health_neglect_2022}. However, the development of robust AI models is hampered by the scarcity of large-scale, expertly annotated datasets, as producing reliable annotations is both costly and time-consuming~\cite{schwendicke2020ai,almalki2023wacv_mim,hu2025ssl_periapical,okazaki2024radimagenet}. 
For example, early caries and subtle periapical changes are visually subtle, low-contrast, and often ambiguous even for expert raters~\cite{mohammadrahimi2022caries,rubak2026labelnoise}. Furthermore, dataset construction and model generalization face the additional hurdle of significant domain shift. 
This shift arises from variability across centers and acquisition protocols, including device-specific differences in sensor properties, exposure, and image compression, all of which can substantially alter image characteristics.
Collectively, these issues of labeled data scarcity and domain shift can lead to severe performance degradation in dental AI models~\cite{zech2018generalization}.

Self-supervised learning (SSL) has emerged as a powerful paradigm for learning transferable visual representations from large-scale unlabeled data, significantly reducing the dependency on costly manual annotations~\cite{huang2023self}. This makes SSL a highly promising direction for developing robust dental AI models. Among various SSL approaches, the DINO framework and its successors are particularly relevant. The original DINO~\cite{caron2021dino} demonstrated that its multi-crop self-distillation strategy could induce emergent object-centric representations in Vision Transformers, enabling a single encoder to support both recognition and dense prediction tasks. Building upon this, DINOv2~\cite{oquab2023dinov2} improved robustness and versatility by scaling up pre-training to 142 million unlabeled images and refining the training recipe, which enhanced its performance across diverse downstream tasks. Most recently, DINOv3~\cite{dinov3} has advanced this lineage by pre-training on an unprecedented scale of 1.7 billion natural images, with an explicit design focus on obtaining high-quality dense token representations. A key innovation, Gram anchoring, effectively mitigates the degradation of patch-level features during prolonged large-scale training, thereby crucially preserving the spatial fidelity of the resulting feature maps. These characteristics---especially the ability to learn dense, spatially precise features---make  DINOv3  a compelling foundation for dental image analysis, where accurately delineating fine anatomical boundaries and detecting subtle pathological changes are of paramount importance.


However, it remains unclear whether representations learned from natural images can reliably transfer to dental radiographs and intraoral photographs, given the challenges of heterogeneous acquisition conditions and low-contrast findings. Even within dental imaging, there are substantial modality gaps: panoramic radiographs suffer from projection overlaps and anatomical clutter, while intraoral photographs exhibit large variations in lighting, specular highlights, and texture~\cite{turosz2023applications,mutawa2026deep}.
In addition, different diagnostic tasks require different levels of fine-grained detail. For example, tooth segmentation requires precise delineation of the outline of each tooth, which requires a model that prioritizes high-resolution spatial details and multi-scale contextual features to preserve boundaries~\cite{hou2023teeth,ma2025dual}. In contrast, caries detection involves both localizing lesions and classifying their severity, which calls for a model capable of integrating region proposal and feature pooling mechanisms.
These considerations raise a central question:
\begingroup
\addtolength\leftmargini{-0.05in}
\begin{quote}
\textit{Can \model{DINOv3} serve as a strong unified encoder for dental image analysis across tasks and modalities?}
\end{quote}
\endgroup

To address this question, we present \model{DinoDental}, a unified benchmark to systematically assess DINOv3 as an off-the-shelf vision encoder for dental image analysis.
We evaluate DINOv3 across multiple public datasets spanning \emph{panoramic radiographs} and \emph{intraoral photographs}, and cover a broad set of task families including \emph{classification}, \emph{detection/instance segmentation}, and \emph{semantic segmentation}.
Crucially, we structure the study from three complementary dimensions:
(i) \textbf{task--modality coverage}, to determine whether DINOv3 can serve as a \emph{unified} encoder across heterogeneous tasks and imaging regimes;
(ii) \textbf{model size and input resolution}, to test whether the common ``bigger model / higher resolution'' intuition remains valid in dentistry or becomes non-monotonic under domain shift;
and (iii) \textbf{adaptation strategy}, to quantify how frozen training, full fine-tuning, and parameter-efficient LoRA~\cite{hu2021lora} trade accuracy for efficiency in realistic multi-task settings.
The main contributions of this paper can be summarized as follows:
\begin{itemize}
\item We introduce DinoDental, a  unified benchmark for evaluating DINOv3's transferability to dental image analysis. It encompasses both panoramic radiographs and intraoral photographs, covering a range of mainstream dental analytical tasks.

\item We conduct a systematic study on how DINOv3's performance is influenced by model size and input image resolution across different tasks. Furthermore, we compare various fine-tuning adaptation strategies to maximize its effectiveness in the dental domain.

\item Extensive experiments on multiple public oral datasets validate DINOv3's strong performance across diverse tasks, demonstrating its significant value and contribution to the dental AI community.



\end{itemize}


\begin{table*}[t]
\caption{Overview of the public datasets included in DinoDental. The benchmark covers panoramic radiographs and intraoral photographs across multiple dental imaging tasks, including classification, detection and instance segmentation, and semantic segmentation, and summarizes the modality, data scale, and task description of each dataset.}
\label{tab:dental_datasets}
\centering
\small
\setlength{\tabcolsep}{6pt}
\renewcommand{\arraystretch}{1.15}
\resizebox{1.0\linewidth}{!}{
\begin{tabular}{
    p{0.20\textwidth}
    p{0.15\textwidth}
    >{\raggedleft\arraybackslash}p{0.16\textwidth}
    p{0.43\textwidth}
}
\toprule
\textbf{Dataset} & \textbf{Modality} & \textbf{Data Scale} & \textbf{Description} \\
\midrule

\multicolumn{4}{c}{\textbf{Object Detection \& Instance Segmentation}} \\
\midrule
OralXrays-9~\cite{chen2025OralXrays} & Panoramic X-ray & 12,688 & Multi-object anomaly detection over 9 categories (84,113 instances) \\
DENTEX~\cite{hamamci2023dentex} & Panoramic X-ray & 1,005 & Abnormal-tooth detection with dental enumeration and 4 diagnosis classes \\
ADCD~\cite{muresanu2024adcd} & Panoramic X-ray & 1,628 & Detection of diverse dental conditions\\
CariesXrays~\cite{chen2024cariesxrays} & Panoramic X-ray & 6,000 & Dental caries detection (13,783 annotated instances) \\
AlphaDent~\cite{sosnin2025alphadent} &  intraoral Photo & 1,320  & Pathology instance segmentation over 9 classes \\
DentalAI~\cite{valluri2023dentalai} &  intraoral Photo & 2,495 & Instance masks over 4 classes (28,904 instances) \\
\midrule

\multicolumn{4}{c}{\textbf{Semantic Segmentation}} \\
\midrule
Multi-center Panoramic~\cite{li2024multicenter} & Panoramic X-ray & 6,536 & Semantic segmentation for impacted teeth / periodontitis / dental caries (multi-center subsets) \\
DC1000~\cite{wang2023multi} & Panoramic X-ray & 1,000 & Caries lesion segmentation with 3 severity levels, more than 7,500 lesions \\
\midrule

\multicolumn{4}{c}{\textbf{Classification}} \\
\midrule
MTMD(single-label)~\cite{mtm_mendeley} & Panoramic X-ray & 1,317 & 4-way mandibular third-molar (wisdom tooth) presence assessment \\
Mixed Dataset (multi-label)~\cite{chen2025OralXrays,muresanu2024adcd,sosnin2025alphadent,valluri2023dentalai} & Panoramic X-ray \& intraoral Photo & 18,131 & Multi-label image classification by aggregating box labels (OralXrays-9, ADCD, AlphaDent, DentalAI) \\
\bottomrule
\end{tabular}}
\end{table*}

\section{Related work}\label{sec:relwork}

\subsection{Self-supervised visual foundation models}
Self-supervised learning (SSL) aims to learn useful visual representations from unlabeled data by constructing supervisory signals directly from the images themselves~\cite{jing2020survey,gui2024survey}. Early vision SSL methods relied on hand-designed pre-text tasks, such as predicting relative patch positions~\cite{doersch2015context}, solving jigsaw-style patch re-ordering problems~\cite{noroozi2016jigsaw,misra2020self}, inpainting missing regions~\cite{pathak2016context}, re-colorizing grayscale images~\cite{zhang2016colorful}, or recognizing synthetic image transformations~\cite{gidaris2018unsupervised}. While these approaches demonstrated that supervision could be derived without manual labels, their transferability was often limited by the narrowness of the designed task.

Modern SSL methods instead learn representations at scale through more general objectives, and can be broadly grouped into three representative families: contrastive learning, self-distillation, and masked image modeling. Contrastive approaches such as SimCLR~\cite{chen2020simclr} and MoCo~\cite{he2020moco} learn instance-discriminative features by aligning different views of the same image while separating different samples in the embedding space. These methods have demonstrated strong transferability, including in medical imaging settings where labeled data are scarce. However, their objectives mainly emphasize image-level semantic invariance, which may limit their advantages for boundary-sensitive dense prediction tasks.

Another important line is self-distillation without explicit negative pairs. BYOL~\cite{grill2020byol} showed that stable representation learning can emerge from asymmetric teacher--student consistency, while DINO~\cite{caron2021dino} further demonstrated that self-distillation with Vision Transformers can induce semantically meaningful and object-aware representations. In particular, DINO revealed that a single visual encoder can support not only image-level recognition but also dense prediction, making the DINO family especially relevant when a unified backbone is desired across multiple downstream tasks.

In parallel, masked image modeling (MIM) methods, including BEiT~\cite{bao2021beit}, MAE~\cite{he2022mae}, and iBOT~\cite{zhou2021ibot}, learn representations through masked prediction or reconstruction and often preserve richer patch-level structure than purely image-level objectives. These characteristics are especially relevant to medical and dental imaging, where subtle lesions, weak boundaries, and fine anatomical details strongly affect downstream performance. Recent studies have increasingly explored SSL and MIM-style pretraining for radiographs and related medical modalities~\cite{hu2025ssl_periapical,okazaki2024radimagenet,cai2024ssad}. Nevertheless, most prior work remains task-specific or modality-specific, and does not examine whether a \emph{single} large visual encoder can support both classification and dense prediction under a unified dental benchmark.

\subsection{Vision--language pretraining for medical image analysis}
Beyond purely visual SSL, vision--language pretraining has emerged as another major direction for transferable medical image representation learning. CLIP~\cite{radford2021clip} established the general paradigm of aligning image and text embeddings at scale, enabling flexible semantic transfer through language supervision. This idea has been extended to the biomedical domain by methods such as ConVIRT~\cite{zhang2022contrastive}, GLoRIA~\cite{huang2021gloria}, MedCLIP~\cite{wang2022medclip}, BioViL~\cite{boecking2022biovil}, and BiomedCLIP~\cite{biomedclip}, which leverage paired reports or medical descriptions to inject clinically meaningful semantics into visual representations.

These studies have shown promise, particularly for recognition and retrieval tasks that benefit from concept-level alignment. However, the effectiveness of vision--language pretraining in medical imaging depends strongly on the availability, quality, and granularity of paired text. Clinical reports may be incomplete, noisy, or focused on only the most salient findings, making it difficult to learn representations for subtle visual abnormalities or fine boundaries. As a result, language-aligned models may be advantageous for semantic recognition, yet offer less benefit for dense localization and delineation tasks. Since our goal is to study a \emph{single} encoder across heterogeneous dental tasks under matched downstream pipelines, we primarily focus on purely visual foundation models while including BiomedCLIP as a representative language-supervised baseline for comparison.

\subsection{DINO family}
DINO performs negative-free cross-view alignment using a teacher--student self-distillation objective:
the student matches the teacher's softened predictions across multi-crop views, while collapse is prevented through output centering and teacher-temperature sharpening~\cite{caron2021dino}. This family is particularly relevant when a \emph{single} encoder must support both classification and dense prediction, as DINO exhibits emergent object-centric behavior in ViTs~\cite{caron2021dino}.
DINOv2 improves transfer robustness by scaling with curated data and a stronger training recipe; crucially for dense tasks, it also strengthens token representations via a \emph{joint} objective that combines image-level DINO self-distillation with an \emph{iBOT-style} patch/token prediction loss, encouraging consistent dense features in addition to global semantics~\cite{oquab2023dinov2}.
DINOv3 further targets localization and segmentation by addressing a dense-prediction failure mode observed in large-scale, long-duration training: even when global embeddings remain strong, dense token features can drift and lose spatial fidelity~\cite{dinov3}.
To mitigate this degradation, it introduces \emph{Gram anchoring} to constrain second-order statistics of dense tokens and explicitly preserve high-quality dense representations for boundary-sensitive tasks, alongside continued scaling and high-resolution, transfer-oriented engineering choices.

These properties are particularly relevant to dental image analysis, where a unified encoder is expected to support both global recognition and fine-grained delineation across heterogeneous modalities, including panoramic radiographs and intraoral photographs. Related segmentation-oriented pipelines also leverage DINO-style dense tokens with task-specific training recipes or architectures, such as SegDINO~\cite{yang2025segdino}. However, recent medical-domain benchmarks suggest that scaling and transfer under domain shift can be non-monotonic and strongly dependent on the target modality and task~\cite{liu2025medical_dinov3_standard}. In dentistry, this question is even less explored: it remains unclear whether a single DINO-family encoder can transfer reliably across classification, detection, instance segmentation, and semantic segmentation under matched downstream protocols. This motivates our controlled benchmark across multiple dental modalities and tasks with fixed heads and decoders.


\section{Benchmark Setup}\label{sec:benchmark}
\subsection{Datasets}\label{sec:datasets}

To rigorously validate DINOv3's generalizability across dental modalities and label granularities, we integrate 10 public datasets of panoramic radiographs and intraoral photographs. Our benchmarks encompass the following tasks:
On Panoramic Radiographs: We benchmark detection and instance segmentation of common findings (e.g., caries, periapical lesions, impacted teeth, restorations). Additionally, we evaluate semantic segmentation using pixel-level masks for structures like impacted teeth, periodontitis regions, and caries areas.
On Intraoral Photographs: We assess detection and instance segmentation of texture-based findings such as caries, cavities, cracks, and other pathologies.
For Image Classification: \model{DinoDental} includes both single-label diagnoses (e.g., mandibular third-molar presence) and multi-label recognition derived from detection annotations. This design allows us to systematically verify the transfer of representations from bounding boxes and instance masks to semantic masks and image-level labels across diverse acquisition conditions.
Detailed descriptions of each dataset are provided in the subsequent sections.

\textbf{Data Splits.} When an official data split is available, we follow the original protocol.
Otherwise, we randomly split the available labeled data into 80\% for training and 20\% for validation.
For datasets that provide only a labeled training split, this partition is performed within that split.

\subsubsection{Object Detection and Instance Segmentation}\label{sec:datasets_det}

\begin{itemize}
    \item \dslink{OralXrays-9}{https://github.com/Binz-Chan/CVPR2025_OralXrays-9}~\cite{chen2025OralXrays}. A hospital-scale benchmark for multi-object oral anomaly detection on panoramic radiographs (CVPR~2025), containing 12,688 images with 84,113 annotated instances across nine anomaly categories: apical periodontitis, dental caries, wisdom tooth, missing tooth, dental filling, root canal filling, dental implant, porcelain crown, and ceramic bridge.

   \item \dslink{DENTEX}{https://dentex.grand-challenge.org/data/}~\cite{hamamci2023dentex}. A data-efficient benchmark on panoramic radiographs introduced at MICCAI~2023 with standardized evaluation under limited supervision. The fully annotated third subset contains 1005 X-rays and is officially split into 705 training images, 50 validation images, and 250 test images. However, only the training split provides publicly available ground-truth annotations, whereas the validation and test labels are withheld. Therefore, in this work, we use only the labeled training portion of the quadrant-enumeration-diagnosis subset for downstream experiments, focusing on detection and instance segmentation with four diagnosis categories: caries, deep caries, periapical lesions, and impacted teeth.

    \item \dslink{ADCD}{https://www.mdpi.com/2075-4418/14/20/2336}~\cite{muresanu2024adcd}.
    A detection dataset covering diverse dental conditions (e.g., caries, periapical lesions, impacted teeth, restorations/devices) on panoramic radiographs.
    The study reports 1,628 annotated images split into 1,375/153/100 for train/val/test, plus an external validation set of 180 multi-center radiographs.

    \item \dslink{CariesXrays}{https://github.com/Binz-Chan/AAAI2024_CariesXrays}~\cite{chen2024cariesxrays}. A caries-focused benchmark on panoramic radiographs, supporting detection/segmentation-style evaluation and complementary to pixel-level caries segmentation datasets.

    \item \dslink{AlphaDent}{https://arxiv.org/abs/2507.22512}~\cite{sosnin2025alphadent}. An intraoral imaging dataset with pathology instance segmentation labels, containing images from 295 patients with 1,320 photographs across nine pathology classes, namely abrasion, filling, crown, caries 1 class, caries 2 class, caries 3 class, caries 4 class, caries 5 class, and caries 6 class.

    \item \dslink{DentalAI}{https://datasetninja.com/dentalai}~\cite{valluri2023dentalai}. An intraoral dataset with pixel-level instance masks.
    It contains 2,495 images with 28,904 labeled instances from four classes (\emph{tooth, caries, cavity, crack}) and provides an official 1,991/254/250 train/val/test split.
\end{itemize}

\subsubsection{Semantic Segmentation}\label{sec:datasets_seg}

\begin{itemize}
   \item \dslink{Multi-center Panoramic Dataset}{https://github.com/qinxin99/qinxini}~\cite{li2024multicenter}. A multi-center, multi-task labeled panoramic radiography dataset covering \emph{impacted teeth, periodontitis, and dental caries}, intended for benchmarking segmentation and classification. It consists of three hospital cohorts with 4,940, 716, and 880 images, totaling 6,536 images, and provides disease-specific subsets including 2,555 impacted-teeth images, 2,735 periodontitis images, and 1,246 dental-caries images. In this work, we only use the training split provided by the dataset.

    \item \dslink{DC1000}{https://github.com/Zzz512/MLUA}~\cite{wang2023multi}.
    A caries semantic segmentation benchmark with 1,000 images and fine-grained lesion annotations across three severity levels, widely used for pixel-level caries delineation.

\end{itemize}
\subsubsection{Classification}\label{sec:datasets_cls}

\begin{itemize}
   \item \dslink{MTMD}{https://data.mendeley.com/datasets/xn5bz6fdm6}~\cite{mtm_mendeley}.
    It provides 1,317 panoramic radiographs labeled for mandibular third molar (wisdom tooth) presence assessment, enabling single-label classification under a four-category protocol: third molar absent on the left but present on the right, third molar present on the left but absent on the right, third molar present on both sides, and third molar absent on both sides.

    \item \dslink{Mixed Dataset}~\cite{chen2025OralXrays,muresanu2024adcd,sosnin2025alphadent,valluri2023dentalai}.
    We derive image-level multi-label targets from \dataset{OralXrays-9}, \dataset{ADCD}, \dataset{AlphaDent}, and \dataset{DentalAI} to construct a multi-label classification dataset.
    This conversion keeps the same category space as detection and simply discards bounding boxes while retaining per-image class presence.

\end{itemize}

\subsection{Metrics}\label{sec:metrics}


\textbf{Detection and instance segmentation.} Our evaluation follows the COCO protocol, which defines Average Precision ($AP$) based on a set of IoU thresholds $\mathcal{T} = \{0.50, 0.55, \ldots, 0.95\}$.
For a fixed threshold $\tau \in \mathcal{T}$ and object category $c$, $AP_c(\tau)$ is the area under the precision-recall curve:
\begin{equation}
AP_c(\tau) = \int_0^1 p_c(r \mid \tau)\,dr,
\end{equation}
where $p_c(r \mid \tau)$ is the precision at recall $r$. The category-averaged result at $\tau$ is denoted as $AP(\tau) = \frac{1}{C}\sum_{c} AP_c(\tau)$.

The final $AP$ (averaged over all categories and thresholds) is:
\begin{equation}
AP = \frac{1}{|\mathcal{T}|} \sum_{\tau \in \mathcal{T}} AP(\tau).
\end{equation}
We report the standard variants: $AP^{b}$ (box detection) and $AP^{m}$ (mask segmentation), along with their strict thresholds $AP^{b}_{50}$, $AP^{b}_{75}$, $AP^{m}_{50}$, and $AP^{m}_{75}$.




\textbf{Semantic segmentation.} We employ standard metrics for binary semantic segmentation (background and target). All metrics are derived from pixel-level counts of true positives ($\mathrm{TP}_c$), false positives ($\mathrm{FP}_c$), and false negatives ($\mathrm{FN}_c$) for each class $c$.

Primary metrics: We report the mean Intersection over Union (mIoU) and the mean Dice coefficient (mDice), both computed via macro-averaging over the $C=2$ classes:
\begin{align}
\mathrm{mIoU} &= \frac{1}{C} \sum_{c=1}^{C} \frac{\mathrm{TP}_c}{\mathrm{TP}_c + \mathrm{FP}_c + \mathrm{FN}_c}, \\
\mathrm{mDice} &= \frac{1}{C} \sum_{c=1}^{C} \frac{2 \cdot \mathrm{TP}_c}{2 \cdot \mathrm{TP}_c + \mathrm{FP}_c + \mathrm{FN}_c}.
\end{align}

Supplementary metrics: We additionally report the macro-averaged Precision and Recall. Their definitions are consistent with those given in the classification section, with the understanding that the underlying contingency counts ($\mathrm{TP}_c, \mathrm{FP}_c, \mathrm{FN}_c$) are accumulated at the pixel level.

\textbf{Classification.}
For single-label classification, let $N$ denote the number of evaluated samples.
For each sample $i \in \{1,\ldots,N\}$, let $y_i$ be the ground-truth label and
$\hat{y}_i$ the predicted label. Let $\mathbb{I}(\cdot)$ be the indicator function, which equals $1$ if the condition is true and $0$ otherwise.
The classification accuracy is defined as:
\begin{equation}
\text{Accuracy}
=
\frac{1}{N}
\sum_{i=1}^{N}
\mathbb{I}(\hat{y}_i = y_i).
\end{equation}

For multi-label classification, we report mean Average Precision (mAP).
For each class $c$, the Average Precision (AP) is computed from its precision--recall curve as:
\begin{equation}
\mathrm{AP}_c
=
\sum_{n=1}^{N_c}
\left(R_{c,n}-R_{c,n-1}\right)P_{c,n},
\end{equation}
where $N_c$ denotes the number of operating points on the precision--recall curve of class $c$, and
$P_{c,n}$ and $R_{c,n}$ denote the precision and recall at the $n$-th operating point, respectively.
The mAP is then obtained by macro-averaging AP over all classes:
\begin{equation}
\mathrm{mAP}
=
\frac{1}{C}\sum_{c=1}^{C}\mathrm{AP}_c,
\end{equation}
where $C$ is the number of classes.
In addition, we also report Precision, Recall, and
F1-score after thresholding prediction scores at $0.5$,
unless stated otherwise.

For both single-label and multi-label settings, let
$\text{TP}$, $\text{FP}$, and $\text{FN}$ denote the numbers of
true positives, false positives, and false negatives for a given class.
The macro-averaged Precision, Recall, and F1-score are defined as:
\begin{equation}
\text{Precision}
=
\frac{1}{C}
\sum_{c=1}^{C}
\frac{\text{TP}}{\text{TP} + \text{FP}},
\end{equation}
\begin{equation}
\text{Recall}
=
\frac{1}{C}
\sum_{c=1}^{C}
\frac{\text{TP}}{\text{TP} + \text{FN}},
\end{equation}
\begin{equation}
\text{F1-score}
=
\frac{1}{C}
\sum_{c=1}^{C}
\frac{2\,\text{TP}}{2\,\text{TP} + \text{FP} + \text{FN}}.
\end{equation}



\subsection{Finetuning Strategies}\label{sec:adapt_strat}
To more comprehensively validate the transferability of DINOv3, we compare three widely used adaptation strategies for transfer learning: frozen-backbone training, full end-to-end fine-tuning, and parameter-efficient LoRA tuning~\cite{houlsby2019adapters,jia2022vpt,chen2022adaptformer,hu2021lora}.

\begin{itemize}
    \item \textbf{Frozen}: the DINOv3 encoder is kept fixed.
    Only lightweight task-specific modules are trained, including projection
    layers, the neck  and the prediction heads.
    \item \textbf{Finetune}: all parameters are updated end-to-end.
    To improve training stability, we apply a smaller learning rate to the
    backbone through a learning-rate multiplier.
    \item \textbf{LoRA}: the backbone remains frozen, while low-rank adapters are injected into attention and MLP layers. We optimize the LoRA parameters jointly with the task-specific modules, keeping the original backbone weights fixed.
\end{itemize}


\subsection{Task Setup}\label{sec:task_adapt}

All experiments are implemented using the OpenMMLab ecosystem.
Specifically, image classification is built on MMPreTrain~\cite{2023mmpretrain},
semantic segmentation on MMSegmentation~\cite{mmseg2020}, and
detection/instance segmentation on MMDetection~\cite{mmdetection},
with unified training and configuration management provided by MMEngine~\cite{mmengine2022}.
We integrate DINOv3 as a plug-and-play backbone by converting its pretrained weights into the OpenMMLab format.
To ensure fair comparisons across backbones, we reuse the standard heads and decoders provided by each framework and keep all task-specific components unchanged.
All experiments are conducted on the NVIDIA A40 platform.

\textbf{Classification.} For the task of classification, we train a linear classifier on the global representation extracted from the last transformer block of DINOv3.
Input images are resized to $512\times512$, and the training batch size is set as 16.
Models are trained for 50 epochs using AdamW with a learning rate of $1\times10^{-4}$ and cosine annealing.


\textbf{Detection and Instance Segmentation.} For detection and instance segmentation, we adopt a standard Mask R-CNN~\cite{he2017maskrcnn}
framework and replace the ResNet backbone~\cite{he2016resnet} with multi-level features extracted
from DINOv3~\cite{dinov3}.
To support different backbone scales, we select four transformer blocks with varying depths.
For a ViT with $L$ blocks, the selected layers correspond to
$\{\lceil L/4\rceil,\,\lceil L/2\rceil,\,\lceil 3L/4\rceil,\,L\}$.
Input images are resized following a single‑scale setting, with the maximum resolution set to 1333×800. A batch size of 4 is used for both training and validation. The models are trained for 12 epochs with the AdamW optimizer, using a learning rate of 3×$10^{-4}$. The learning rate schedule consists of a linear warmup during the first epoch, followed by cosine annealing thereafter. 



\textbf{Semantic Segmentation.} For semantic segmentation, we employ an encoder-decoder architecture equipped with a feature pyramid network (FPN)~\cite{lin2017fpn}  neck and a UPerNet-style decoder~\cite{xiao2018upernet}, along with an auxiliary FCN head. During training, inputs are randomly cropped to $1024\times1024$ and augmented with random resizing (in the range of 0.5--2.0), random cropping, and grayscale-preserving intensity augmentation. We use a batch size of 4 for training. 
The model is trained for 20k iterations using the AdamW optimizer with an initial learning rate of $3\times10^{-4}$ and a Poly learning rate scheduler. For fine-tuning, a learning rate multiplier of 0.1 is applied to the backbone.

\section{Experiments}\label{sec:experiments}


We evaluate DINOv3 as a unified vision encoder for dental image analysis from three complementary dimensions:
(i) \textbf{task-modality adaptability}, to determine whether DINOv3 can serve as a \emph{unified} encoder across heterogeneous tasks and dental modalities;
(ii) \textbf{model size and input resolution}, to test whether the common ``bigger model / higher resolution'' intuition remains valid in dentistry;
and (iii) \textbf{adaptation strategy}, to quantify how frozen training, full fine-tuning, and parameter-efficient LoRA trade accuracy and efficiency in realistic dental tasks.
Unless otherwise specified, we follow the unified protocols described in Sec.~\ref{sec:task_adapt}.

\subsection{Task-modality Adaptability}\label{sec:exp_unified}

\subsubsection{Detection and Instance Segmentation}
We begin with the task of detection and instance segmentation, as these tasks directly assess a unified encoder's capability for fine-grained localization and boundary-aware understanding across diverse dental modalities. Performance is evaluated separately on panoramic radiographs (Table~\ref{tab:maskrcnn_12ep_4datasets_ap_bm}) and intraoral photographs (Table~\ref{tab:maskrcnn_12ep_2datasets_ap_bm_singlebest}).
Overall, the results reveal a clear modality-dependent pattern. On panoramic radiographs, no single backbone consistently dominates across all datasets, indicating dataset-specific optimal choices. In contrast, for intraoral photographs, DINOv3 demonstrates a clearer and more consistent advantage, particularly under stricter localization criteria.

\textbf{Panoramic Radiographs.} Performance on the four panoramic benchmarks shows that the optimal backbone varies with dataset characteristics. On \dataset{DENTEX}, DINOv3-L achieves the best performance, while on \dataset{OralXrays-9} it remains competitive but does not surpass the strongest supervised baseline. On \dataset{DENTEX}, it attains $AP^b=34.40\%$ and $AP^b_{75}=38.50\%$, outperforming DINOv2-L by 2.3 and 5.3 points, respectively, and surpassing BiomedCLIP by more than 11 points in box AP. This highlights DINOv3's enhanced capability for precise localization in settings with limited or varied supervision. On \dataset{ADCD} and \dataset{CariesXrays}, which involve smaller and more subtle pathological targets, classic CNN backbones remain highly competitive. ResNet-50 achieves the best $AP^b$ scores of 19.60\% and 46.80\%, respectively, suggesting that traditional detector inductive biases and local feature aggregation are still effective for these specific tasks. On the larger-scale \dataset{OralXrays-9}, performance differences among stronger backbones diminish, indicating that backbone choice becomes less critical when both data scale and supervision are sufficient.

\textbf{Intraoral Photographs.} A more consistent advantage for DINOv3 is observed on intraoral images, which present complex textures, specular highlights, and ambiguous boundaries.
On \dataset{DentalAI}, DINOv3-L achieves the best performance with $AP^b=36.70\%$ and $AP^m=35.80\%$, outperforming DINOv2-L by 2.1 and 2.4 points, respectively. Compared with BEiT-B, the gain exceeds 11 points in box AP and 9 points in mask AP. Similarly, on \dataset{AlphaDent}, DINOv3-L leads with $AP^b=26.5\%$ and $AP^m=25.3\%$, exceeding DINOv2-L by 1.2 and 1.3 points, respectively. These gains are most pronounced at stricter IoU thresholds (e.g., $AP^m_{75}$), underscoring DINOv3's strength in fine-grained boundary delineation for challenging real-world oral scenes.

\begin{figure*}[t]
    \centering
    \includegraphics[width=1\linewidth]{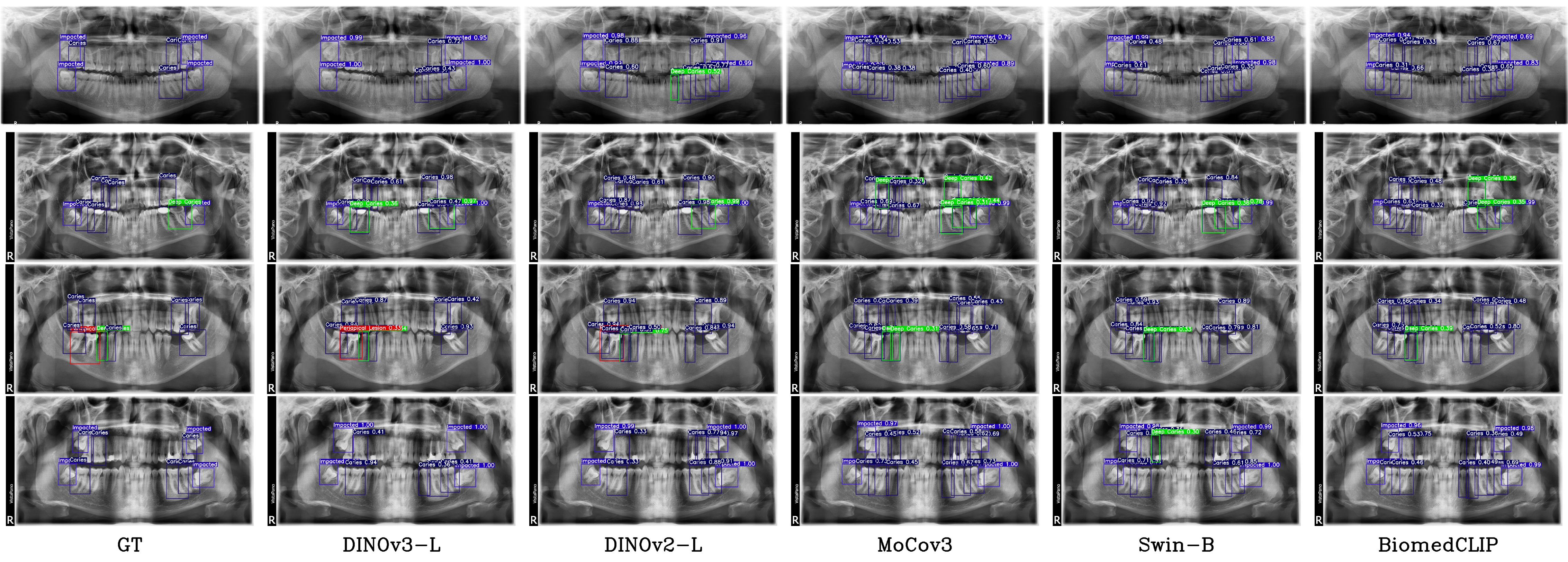}
    \caption{Qualitative detection comparisons on the DENTEX dataset.
    }
    
    \label{fig:dentex_vis}
\end{figure*}

\begin{figure*}[t]
    \centering
    \includegraphics[width=1.0\linewidth]{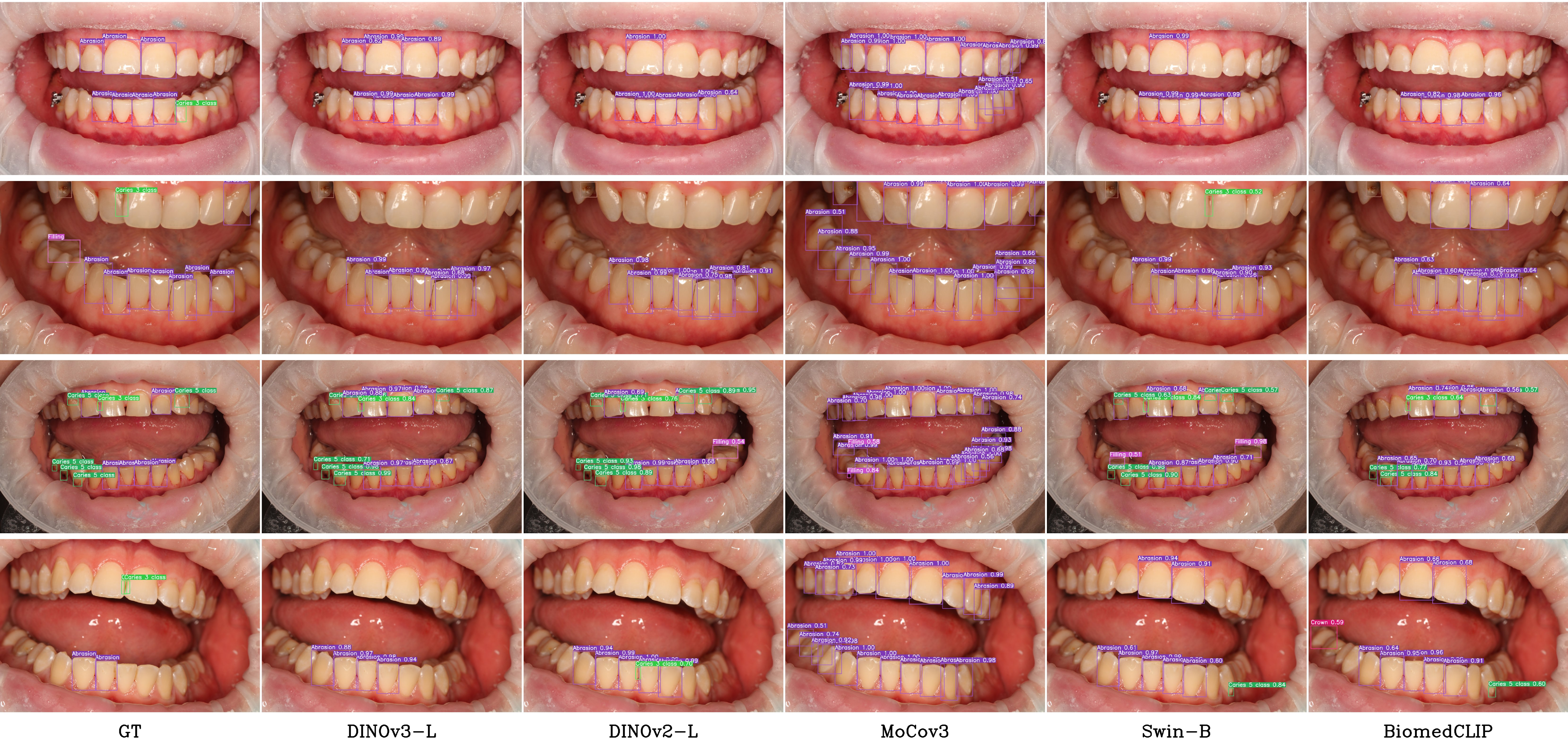}
    \caption{Qualitative detection comparisons on the AlphaDent dataset.
    }
    \label{fig:alphadent_vis}
\end{figure*}

In Figure~\ref{fig:dentex_vis} and Figure~\ref{fig:alphadent_vis}, we visualize the detection results of different backbones on panoramic radiographs from \dataset{DENTEX} and intraoral photographs from \dataset{AlphaDent}, respectively. On \dataset{DENTEX}, DINOv3-L produces detection boxes that are overall closer to the ground truth, with more accurate localization of impacted teeth and caries-related regions and fewer redundant predictions in crowded dental areas. Compared with DINOv3-L, DINOv2-L more often shows slight box misalignment or incomplete retrieval of difficult targets, especially for subtle abnormalities and less prominent lesion regions. MoCov3 exhibits the most obvious over-detection behavior, generating many redundant boxes and spurious responses around overlapping tooth structures. Swin-B and BiomedCLIP are relatively more conservative and therefore miss target instances or fail to fully retrieve some difficult lesion regions in several cases. On \dataset{AlphaDent}, DINOv3-L likewise yields tighter and more coherent object-centric predictions under challenging illumination and complex texture variation. Its predicted boxes better match the GT around subtle abnormal regions, whereas DINOv2-L more often produces slightly shifted boxes or incomplete detections, MoCov3 generates overly dense responses with unnecessary duplicate boxes, and Swin-B and BiomedCLIP miss some small or weakly visible lesion regions.This may be attributed to the stronger dense visual representations learned by DINOv3-L, which better encode tooth structures and lesion-related cues across both panoramic and intraoral images, allowing the detector to distinguish subtle abnormalities and localize them more precisely.

In summary, DINOv3 emerges as a robust and reliable backbone for dental detection and instance segmentation. While no single model dominates every panoramic benchmark, DINOv3 consistently excels in scenarios demanding precise spatial reasoning, such as the \dataset{DENTEX} dataset or texture-rich intraoral photographs. Its advantages are particularly evident under stricter localization criteria, suggesting its dense token representations are highly beneficial for fine-grained dental image analysis.

\begin{TabTwo}[t]

\caption{
Performance comparison of supervised and self-supervised methods on detection and instance segmentation tasks of panoramic X-ray datasets.
}

\label{tab:maskrcnn_12ep_4datasets_ap_bm}

\begin{adjustbox}{max width=\linewidth}
\begin{tabular}{l cccccc cccccc cccccc cccccc}
\toprule
\textbf{Backbone} &
\multicolumn{6}{c}{\textbf{DENTEX}} &
\multicolumn{6}{c}{\textbf{ADCD}} &
\multicolumn{6}{c}{\textbf{CariesXrays}} &
\multicolumn{6}{c}{\textbf{OralXrays-9}} \\
\cmidrule(lr){2-7}\cmidrule(lr){8-13}\cmidrule(lr){14-19}\cmidrule(lr){20-25}
& $AP^{b}$ & $AP^{b}_{50}$ & $AP^{b}_{75}$ & $AP^{m}$ & $AP^{m}_{50}$ & $AP^{m}_{75}$
& $AP^{b}$ & $AP^{b}_{50}$ & $AP^{b}_{75}$ & $AP^{m}$ & $AP^{m}_{50}$ & $AP^{m}_{75}$
& $AP^{b}$ & $AP^{b}_{50}$ & $AP^{b}_{75}$ & $AP^{m}$ & $AP^{m}_{50}$ & $AP^{m}_{75}$
& $AP^{b}$ & $AP^{b}_{50}$ & $AP^{b}_{75}$ & $AP^{m}$ & $AP^{m}_{50}$ & $AP^{m}_{75}$ \\
\midrule

\multicolumn{25}{c}{\textbf{Supervised methods}} \\
\midrule

\bk{ResNet-50}{he2016resnet}      & 26.40 & 42.40 & 28.90 & 26.50 & 42.70 & 32.40
& \best{19.60} & \best{46.50} & 12.40 & \best{19.60} & \best{46.60} & \best{13.10}
& \best{46.80} & \best{82.20} & 48.30 & \best{21.80} & \best{40.40} & \best{21.40}
& \best{68.40} & \best{91.30} & \best{72.30} & \best{68.00} & \best{91.40} & 72.20 \\

\bk{ConvNeXt-B}{liu2022convnext}    & 30.60 & 50.20 & 33.80 & 31.60 & 50.30 & 38.60
& 19.40 & 46.20 & \best{12.90} & 19.20 & 45.80 & 12.70
& 46.00 & 80.40 & \best{48.40} & 21.10 & 39.30 & 21.00
& 67.20 & 90.10 & 71.60 & 67.10 & 90.10 & 71.60 \\

\bk{Swin-B}{liu2021swin}   & 32.00 & 52.80 & 35.50 & 32.50 & 52.40 & 38.20
& 19.40 & 46.30 & 11.70 & 18.90 & 45.20 & 10.40
& 43.00 & 77.80 & 43.80 & 19.90 & 38.30 & 19.10
& 66.40 & 90.30 & 71.30 & 66.10 & 90.30 & 70.90 \\

\bk{DeiT-B}{touvron2021deit}         & 21.60 & 37.70 & 21.30 & 22.00 & 37.40 & 24.80
& 17.20 & 43.10 & 9.80 & 17.10 & 43.10 & 9.50
& 41.30 & 77.20 & 40.20 & 15.10 & 31.70 & 12.30
& 67.40 & 90.80 & 71.70 & 67.20 & 90.70 & 71.80 \\

\bk{AugReg-L}{steiner2022train_vit}       & 21.80 & 38.20 & 22.20 & 22.00 & 38.30 & 23.40
& 17.10 & 42.20 & 10.70 & 16.60 & 41.50 & 9.70
& 39.80 & 76.30 & 37.70 & 12.60 & 28.10 & 9.60
& 66.60 & 90.40 & 71.60 & 66.40 & 90.50 & 71.50 \\

\midrule
\multicolumn{25}{c}{\textbf{Self-supervised methods}} \\
\midrule

\bk{MAE-B}{he2022mae}         & 22.00 & 36.60 & 23.90 & 21.90 & 36.60 & 25.60
& 17.30 & 42.30 & 10.70 & 17.90 & 43.50 & 11.00
& 42.80 & 78.70 & 42.30 & 20.00 & 38.70 & 18.20
& 65.30 & 90.50 & 71.30 & 64.90 & 90.60 & 71.20 \\

\bk{MoCov3}{chen2021mocov3}         & 28.90 & 48.20 & 31.30 & 29.90 & 46.50 & 35.40
& 19.30 & 45.40 & 12.40 & 19.30 & 45.90 & 12.30
& 44.90 & 81.80 & 44.10 & 20.60 & 40.00 & 18.90
& 67.60 & 90.80 & \best{72.30} & 67.20 & 91.00 & \best{72.60} \\

\bk{BiomedCLIP}{biomedclip}     & 23.10 & 39.90 & 24.70 & 23.80 & 40.00 & 26.90
& 17.30 & 43.10 & 9.40 & 17.40 & 42.90 & 9.10
& 37.00 & 73.00 & 34.00 & 17.10 & 35.10 & 14.40
& 63.60 & 89.10 & 69.40 & 63.00 & 89.20 & 69.00 \\

\bk{UniPerceiver-B}{zhu2021uniperceiver} & 24.90 & 43.30 & 25.00 & 24.80 & 43.20 & 27.60
& 18.30 & 44.60 & 11.90 & 18.30 & 45.40 & 11.30
& 42.20 & 78.60 & 42.00 & 14.90 & 31.30 & 12.60
& 67.60 & 90.40 & 72.10 & 67.50 & 90.60 & 72.00 \\

\bk{BEiT-B}{bao2021beit}         & 19.20 & 35.30 & 18.80 & 18.10 & 35.60 & 15.80
& 15.30 & 37.40 & 9.30 & 15.10 & 37.70 & 9.00
& 40.20 & 76.60 & 37.50 & 13.50 & 28.80 & 10.40
& 66.30 & 90.10 & 71.40 & 66.10 & 90.30 & 71.00 \\

\bk{DINOv2-S}{oquab2023dinov2}     & 27.30 & 47.70 & 27.00 & 28.10 & 46.50 & 32.20
& 18.50 & 45.00 & 10.90 & 18.70 & 45.10 & 10.10
& 39.40 & 75.40 & 38.10 & 18.50 & 37.20 & 16.80
& 65.60 & 89.50 & 71.00 & 65.50 & 89.50 & 70.70 \\

\bk{DINOv2-B}{oquab2023dinov2}     & 28.70 & 48.70 & 29.40 & 29.20 & 48.50 & 32.70
& 18.80 & 44.70 & 12.30 & 18.40 & 44.50 & 12.20
& 40.80 & 76.90 & 40.20 & 19.00 & 37.80 & 17.70
& 65.70 & 89.20 & 70.50 & 65.60 & 89.40 & 70.50 \\

\bk{DINOv2-L}{oquab2023dinov2}    & 32.10 & 53.70 & 33.20 & 34.10 & 53.30 & \best{41.50}
& 18.60 & 45.40 & 11.50 & 18.40 & 45.30 & 11.40
& 42.10 & 77.60 & 42.40 & 19.70 & 38.30 & 18.40
& 66.00 & 89.30 & 71.30 & 65.80 & 89.50 & 70.90 \\

\bk{DINOv3-S}{dinov3}    & 30.60 & 48.80 & 36.10 & 31.30 & 48.60 & 38.00
& 18.70 & 44.50 & 11.80 & 18.90 & 44.90 & 12.00
& 40.80 & 77.00 & 39.70 & 18.80 & 37.80 & 17.00
& 66.10 & 90.30 & 71.30 & 66.20 & 90.40 & 71.30 \\

\bk{DINOv3-B}{dinov3}    & 31.80 & 51.40 & 35.50 & 32.60 & 50.80 & 38.60
& 19.20 & 45.10 & 12.60 & 19.40 & 45.20 & 12.00
& 41.90 & 79.00 & 40.70 & 19.70 & 38.80 & 18.10
& 65.90 & 89.80 & 71.30 & 65.70 & 89.80 & 71.10 \\

\bk{DINOv3-L}{dinov3}     & \best{34.40} & \best{54.70} & \best{38.50} & \best{35.60} & \best{55.30} & 41.10
& 19.00 & 44.30 & 12.70 & 18.80 & 43.60 & 13.00
& 43.30 & 78.90 & 44.70 & 20.40 & 38.70 & 20.40
& 65.80 & 89.30 & 70.80 & 65.60 & 89.40 & 70.60 \\
\bottomrule
\end{tabular}%
\end{adjustbox}
\end{TabTwo}

\begin{TabTwo}[t]

\caption{
Performance comparison of supervised and self-supervised methods on detection and instance segmentation tasks of intraoral datasets.
}

\label{tab:maskrcnn_12ep_2datasets_ap_bm_singlebest}

\begin{adjustbox}{max width=\linewidth}
\begin{tabular}{l cccccc cccccc}
\toprule
\textbf{Backbone} &
\multicolumn{6}{c}{\textbf{DentalAI}} &
\multicolumn{6}{c}{\textbf{AlphaDent}} \\
\cmidrule(lr){2-7}\cmidrule(lr){8-13}
& $AP^{b}$ & $AP^{b}_{50}$ & $AP^{b}_{75}$ & $AP^{m}$ & $AP^{m}_{50}$ & $AP^{m}_{75}$
& $AP^{b}$ & $AP^{b}_{50}$ & $AP^{b}_{75}$ & $AP^{m}$ & $AP^{m}_{50}$ & $AP^{m}_{75}$ \\
\midrule

\multicolumn{13}{c}{\textbf{Supervised methods}} \\
\midrule

\bk{ResNet-50}{he2016resnet}      & 30.00 & 47.10 & 30.80 & 30.50 & 46.80 & 31.30 & 22.20 & 38.70 & 21.10 & 21.30 & 36.70 & 19.70 \\
\bk{Swin-B}{liu2021swin}         & 33.20 & 52.60 & 32.70 & 33.20 & 51.90 & 34.10 & 24.60 & 44.30 & 21.10 & 22.70 & 40.20 & 21.30 \\
\bk{ConvNeXt-B}{liu2022convnext}     & 33.80 & 54.90 & 32.50 & 33.40 & 52.30 & 33.70 & 23.40 & 42.70 & 21.30 & 22.20 & 38.00 & 20.30 \\
\bk{AugReg-L}{steiner2022train_vit}       & 30.50 & 50.60 & 29.70 & 30.50 & 47.90 & 30.90 & 21.00 & 38.90 & 19.60 & 20.30 & 35.60 & 18.00 \\
\bk{DeiT-B}{touvron2021deit}         & 31.70 & 50.40 & 31.30 & 31.40 & 49.30 & 32.00 & 22.50 & 41.40 & 20.80 & 21.60 & 37.90 & 19.80 \\

\midrule
\multicolumn{13}{c}{\textbf{Self-supervised methods}} \\
\midrule

\bk{MAE-B}{he2022mae}         & 29.90 & 47.10 & 30.10 & 30.70 & 46.50 & 31.90 & 20.10 & 36.00 & 17.60 & 19.40 & 32.50 & 18.10 \\
\bk{MoCov3}{chen2021mocov3}         & 32.00 & 50.50 & 32.10 & 32.00 & 49.10 & 32.20 & 23.20 & 42.10 & 20.80 & 22.50 & 39.60 & 21.40 \\
\bk{UniPerceiver-B}{zhu2021uniperceiver} & 32.10 & 51.70 & 30.90 & 31.30 & 49.50 & 31.60 & 23.10 & 41.70 & 21.70 & 22.20 & 39.40 & 21.30 \\
\bk{BEiT-B}{bao2021beit}         & 25.60 & 43.30 & 25.70 & 26.40 & 41.30 & 27.30 & 15.40 & 28.10 & 14.60 & 14.60 & 26.00 & 13.20 \\
\bk{BiomedCLIP}{biomedclip}     & 28.00 & 45.60 & 28.70 & 28.40 & 44.40 & 29.30 & 18.70 & 33.70 & 17.80 & 18.40 & 31.40 & 17.80 \\
\bk{DINOv2-S}{oquab2023dinov2}       & 32.60 & 53.50 & 31.30 & 32.30 & 51.30 & 33.60 & 24.80 & 44.40 & 22.50 & 23.30 & 40.80 & 21.80 \\
\bk{DINOv2-B}{oquab2023dinov2}      & 33.20 & 52.50 & 33.30 & 33.00 & 52.10 & 32.90 & 26.00 & 45.40 & 23.60 & 25.30 & 42.40 & 23.00 \\
\bk{DINOv2-L}{oquab2023dinov2}       & 34.60 & 53.00 & 34.80 & 33.40 & 52.30 & 33.40 & 25.30 & 42.80 & 23.90 & 24.00 & 40.60 & 22.20 \\
\bk{DINOv3-S}{dinov3}       & 33.90 & 53.70 & 33.40 & 33.60 & 51.50 & 33.40 & 25.00 & 45.70 & 22.80 & 24.40 & \best{43.00} & 23.40 \\
\bk{DINOv3-B}{dinov3}       & 35.70 & 55.30 & 35.80 & 34.50 & 51.90 & 35.10 & 25.50 & 43.80 & 23.80 & 24.60 & 42.60 & 22.90 \\
\bk{DINOv3-L}{dinov3}       & \best{36.70} & \best{57.80} & \best{38.50} & \best{35.80} & \best{55.70} & \best{35.50}
              & \best{26.50} & \best{46.70} & \best{25.10} & \best{25.30} & 42.80 & \best{23.50} \\
\bottomrule
\end{tabular}%
\end{adjustbox}
\end{TabTwo}

\subsubsection{Semantic Segmentation}\label{sec:exp_seg}

In Table~\ref{tab:uperhead_20k_3datasets_miou_mdice_pr_re}, we evaluate the performance of DINOv3 backbones within the UPerNet framework for panoramic semantic segmentation across three dental disease datasets: \textit{Impacted}, \textit{Periodontitis}, and \textit{Caries}. We can observe from the table that DINOv3 variants, especially the large model (DINOv3-L), demonstrate highly competitive performance across all datasets. Notably, DINOv3 achieves near-best performance on the \textit{Periodontitis} dataset, remaining highly competitive with the strongest supervised backbones.
For the \textit{Impacted} dataset, DINOv3-L achieves an mIoU of 89.04\%, which is comparable to the top-performing supervised backbones like ConvNeXt-B (89.77\%) and ResNeSt-101 (89.81\%). 
It outperforms other self-supervised models such as MAE-B (87.43\%) and BEiT-B (88.84\%). This indicates that DINOv3's self-supervised features effectively transfer to the task of impacted tooth segmentation, nearly closing the gap with supervised pretraining.
DINOv3-L delivers outstanding performance on \textit{Periodontitis}, achieving an mIoU of 96.32\%, which is essentially on par with the best-performing backbones. This surpasses the strong supervised baseline ConvNeXt-B (96.28\%) and significantly exceeds other self-supervised methods like MAE-B (87.61\%) and BEiT-B (88.57\%). The high mDice (98.10\%), Precision (98.22\%), and Recall (97.99\%) scores further confirm the model's robustness and accuracy in segmenting periodontitis regions, suggesting its features are particularly well-suited for this pathology.
On the \textit{Caries} dataset, DINOv3-L attains an mIoU of 78.66, remaining competitive among self-supervised methods. While slightly lower than the top supervised model ConvNeXt-B (79.15\%), it shows a clear advantage over other self-supervised approaches like MAE-B (75.24\%) and is competitive with BEiT-B (78.59\%). The balanced Precision (89.97\%) and Recall (83.62\%) indicate reliable detection without significant bias towards false positives or negatives.

A clear scaling trend is observed within the DINOv3 family: performance generally improves from the small (S) to the large (L) variant across most metrics and datasets. For instance, on \textit{Periodontitis}, mIoU increases from 96.02\%(S) to 96.32\%(L). This trend validates the benefit of larger model capacity within the DINOv3 self-supervised paradigm for medical image segmentation.
These results demonstrate that features learned via DINOv3's self-supervised paradigm transfer effectively to diverse dental disease segmentation tasks, offering a powerful alternative to supervised pretraining, particularly in scenarios where labeled medical data is scarce.

\begin{figure*}[t]
    \centering
    \includegraphics[width=1\linewidth]{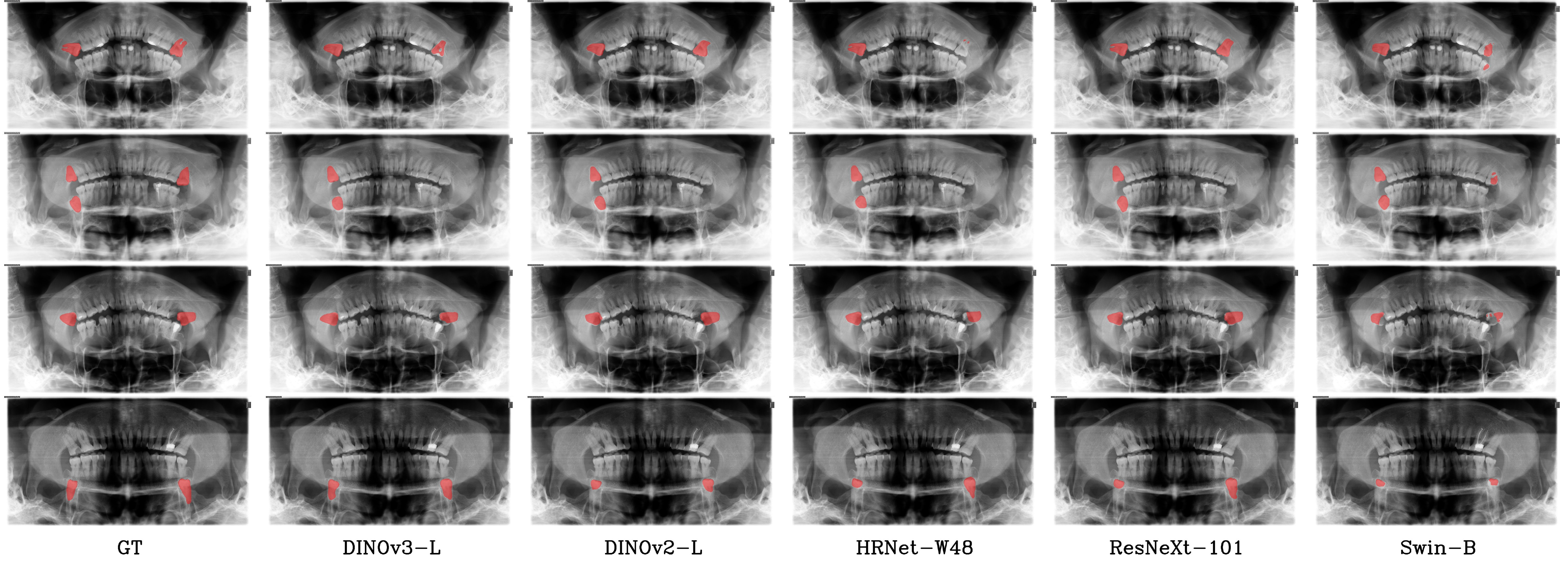}
    \caption{Qualitative segmentation comparison on the Multi-center Panoramic dataset.
    Red masks indicate impacted-tooth regions. 
    }
    \label{fig:impacted_seg_vis}
\end{figure*}

In Figure~\ref{fig:impacted_seg_vis}, we visualize the segmentation results of different backbones on the \dataset{Impacted} dataset. 
Compared with the other methods, DINOv3-L produces segmentation masks that are overall closer to the ground truth, with more complete target coverage and more regular contours. In particular, its predictions better preserve the full extent of the impacted regions and maintain clearer anatomical shapes, whereas DINOv2-L, HRNet-W48, and ResNeXt-101 more often show boundary shrinkage or partial omission around the lesion areas. Swin-B exhibits less stable predictions, with more fragmented responses and small spurious activations in ambiguous regions. By contrast, DINOv3-L suppresses these false-positive fragments more effectively while retaining the main lesion structures, leading to masks that are visually cleaner and more contiguous. This advantage is especially evident in relatively small or slender impacted regions, where competing methods tend to under-segment the target or miss part of its contour, while DINOv3-L remains more faithful to the GT. This improved delineation can be attributed to the stronger dense token representations of DINOv3-L, which better encode boundary and regional information in panoramic radiographs, allowing the model to capture subtle structural variations and recover more complete lesion shapes.

\begin{TabTwo}[t]
\caption{Performance comparison of supervised and self-supervised methods on semantic segmentation tasks of panoramic X-ray datasets.}

\label{tab:uperhead_20k_3datasets_miou_mdice_pr_re}

\begin{adjustbox}{max width=\linewidth}
\begin{tabular}{l cccc cccc cccc}
\toprule
\textbf{Backbone} &
\multicolumn{4}{c}{\textbf{Impacted}} &
\multicolumn{4}{c}{\textbf{Periodontitis}} &
\multicolumn{4}{c}{\textbf{Caries}} \\
\cmidrule(lr){2-5}\cmidrule(lr){6-9}\cmidrule(lr){10-13}
&
\textbf{mIoU}  & \textbf{mDice}  & \textbf{Prec.}  & \textbf{Rec.}  &
\textbf{mIoU}  & \textbf{mDice}  & \textbf{Prec.}  & \textbf{Rec.}  &
\textbf{mIoU}  & \textbf{mDice}  & \textbf{Prec.}  & \textbf{Rec.}  \\
\midrule

\multicolumn{13}{c}{\textbf{Supervised methods}} \\
\midrule

\bk{ViT-B}{dosovitskiy2021vit}          & 88.03 & 93.22 & 92.68 & 93.77  & 88.01 & 93.21 & 93.73 & 92.70  & 76.92 & 85.10 & 88.74 & 82.10 \\
\bk{Twins-S}{chu2021twins}        & 87.10 & 92.62 & 92.12 & 93.13  & 94.65 & 97.20 & 97.81 & 96.61  & 77.95 & 85.94 & 89.06 & 83.30 \\
\bk{Swin-B}{liu2021swin}         & 81.59 & 88.77 & 91.35 & 86.50  & 95.91 & 97.88 & 98.01 & 97.75  & 70.00 & 78.76 & 88.54 & 73.00 \\
\bk{ResNeXt-101}{xie2017resnext}    & 89.39 & 94.08 & 94.01 & 94.15  & 95.71 & 97.78 & 97.98 & 97.57  & 77.67 & 85.72 & 89.91 & 82.34 \\
\bk{ResNet-50}{he2016resnet}      & 86.54 & 92.25 & 93.17 & 91.37  & 94.90 & 97.34 & 97.89 & 96.81  & 75.25 & 83.67 & 90.42 & 78.89 \\
\bk{ResNet-101}{he2016resnet}     & 89.70 & 94.27 & 93.97 & 94.57  & 95.66 & 97.75 & 98.37 & 97.14  & 77.93 & 85.93 & 90.43 & 82.34 \\
\bk{ResNeSt-101}{zhang2022resnest}  & \best{89.81} & \best{94.34} & 94.13 & 94.56  & 95.51 & 97.67 & 97.90 & 97.43   & 77.81 & 85.83 & 91.52 & 81.53 \\
\bk{PoolFormer-S12}{yu2022poolformer} & 89.77 & 94.31 & 93.75 & \best{94.89}  & \best{96.33} & \best{98.11} & 98.06 & \best{98.15}  & 77.44 & 85.52 & 90.24 & 81.81 \\
\bk{MSCAN-T}{guo2022segnext}        & 87.34 & 92.78 & 93.73 & 91.86  & 94.80 & 97.28 & 97.78 & 96.80  & 76.63 & 84.85 & 90.53 & 80.59 \\
\bk{HRNet-W48}{wang2019hrnet}      & 87.89 & 93.13 & 93.98 & 92.31  & 94.87 & 97.32 & 97.47 & 97.18  & 78.05 & 86.02 & 90.97 & 82.16 \\
\bk{ConvNeXt-B}{liu2022convnext}     & 89.77 & 94.31 & 94.09 & 94.55  & 96.28 & 98.08 & 98.15 & 98.01  & 79.15 & 86.91 & 89.85 & 84.38 \\

\midrule
\multicolumn{13}{c}{\textbf{Self-supervised methods}} \\
\midrule

\bk{MAE-B}{he2022mae}       & 87.43 & 92.83 & 93.08 & 92.59  & 87.61 & 92.95 & 93.28 & 92.62  & 75.24 & 83.66 & 91.31 & 78.43 \\
\bk{BEiT-B}{bao2021beit}         & 88.84 & 93.73 & 93.93 & 93.54  & 88.57 & 93.56 & 94.05 & 93.09  & 78.59 & 86.46 & \best{92.26} & 82.08 \\
\bk{DINOv2-S}{oquab2023dinov2}      & 87.39 & 92.80 & 94.51 & 91.23  & 96.07 & 97.97 & 98.36 & 97.59  & 78.39 & 86.30 & 91.06 & 82.54 \\
\bk{DINOv2-B}{oquab2023dinov2}       & 87.73 & 93.02 & \best{94.56} & 91.59  & 96.13 & 98.00 & 98.30 & 97.70  & \best{79.58} & \best{87.24} & 90.62 & \best{84.39} \\
\bk{DINOv2-L}{oquab2023dinov2}       & 87.42 & 92.82 & 93.49 & 92.18  & 95.70 & 97.77 & \best{98.40} & 97.16  & 77.94 & 85.94 & 90.25 & 82.48 \\
\bk{DINOv3-S}{dinov3}       & 87.90 & 93.14 & 93.43 & 92.85  & 96.02 & 97.94 & 97.93 & 97.96  & 76.56 & 84.80 & 90.22 & 80.69 \\
\bk{DINOv3-B}{dinov3}       & 88.68 & 93.63 & 92.64 & 94.67  & 96.22 & 98.05 & 98.08 & 98.02  & 77.37 & 85.47 & 89.26 & 82.36 \\
\bk{DINOv3-L}{dinov3}       & 89.04 & 93.86 & 93.09 & 94.66  & 96.32 & 98.10 & 98.22 & 97.99  & 78.66 & 86.52 & 89.97 & 83.62 \\
\bottomrule
\end{tabular}
\end{adjustbox}
\end{TabTwo}

\subsubsection{Classification}\label{sec:exp_cls}

In Table~\ref{tab:cls_3settings_bigtable_50ep} and Table~\ref{tab:ioscan_multilabel_bigtable_50ep}, we turn to classification tasks to validate the performance of DINOv3 in robust global semantic recognition.

On panoramic radiographs (Table~\ref{tab:cls_3settings_bigtable_50ep}), DINOv3 performs strongly across all three benchmarks. Specifically, on \dataset{OralXrays-9}, DINOv3-B reaches 95.49\% mAP, exceeding the strongest supervised baseline Swin-B(94.99\%) by 0.5 points and DINOv2-B by 0.22 points, while DINOv3-S achieves the best F1 score of 85.95\%. On \dataset{ADCD}, DINOv3-S obtains the best mAP of 73.06\%, outperforming the strongest supervised baseline RegNetX-8GF by 4.34 points, and DINOv3-B achieves the best F1 score of 64.43\%. On the single-label \dataset{MTMD} benchmark, DINOv2-S attains the highest accuracy, but DINOv3-L achieves the best F1 score, suggesting a better balance across categories even when the top-1 accuracy is not maximal.
On intraoral photographs (Table~\ref{tab:ioscan_multilabel_bigtable_50ep}), DINOv3 is particularly strong on precision-sensitive metrics. DINOv3-L achieves the best performance on \dataset{DentalAI} in both precision and F1 score, surpassing DINOv2-L by 4.03 and 2.68 points, respectively. This indicates that DINOv3 is especially effective in reducing false positives while preserving strong recognition ability in texture-rich intraoral scenes. On \dataset{AlphaDent}, DINOv2-L remains the strongest model, achieving the best mAP and F1 score, while DINOv3-L stays competitive without any dataset-specific backbone selection or tuning.

\begin{figure*}[t]
    \centering
    \includegraphics[width=0.98\linewidth]{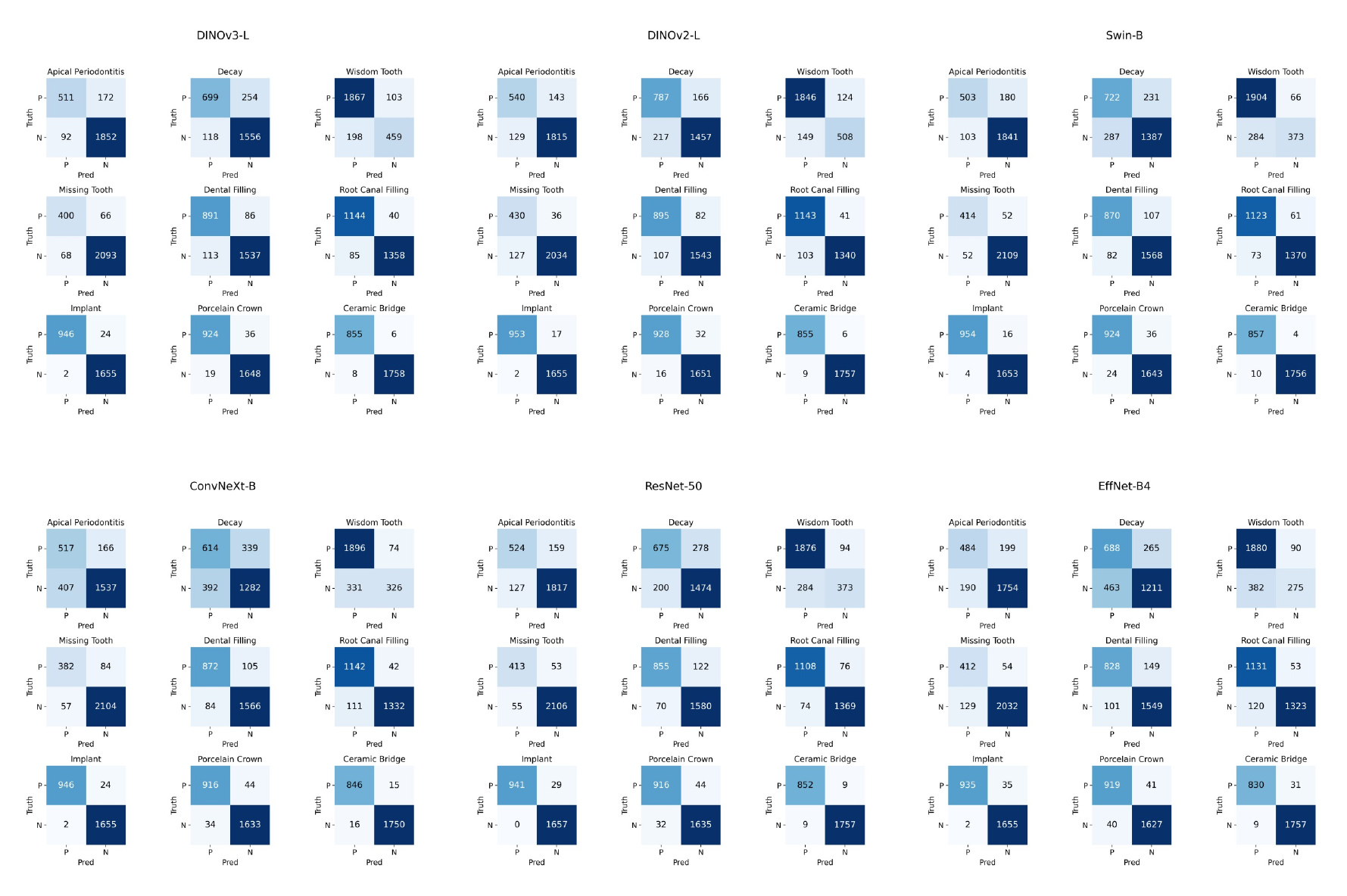}
    \caption{Confusion matrices of the multi-label classification on OralXrays-9 dataset.
    }
    \label{fig:oralx9_confmat}
\end{figure*}

In Figure~\ref{fig:oralx9_confmat}, we visualize the class-wise confusion matrices of different backbones to further examine the multi-label classification performance on \dataset{OralXrays-9}. Compared with the other backbones, DINOv3-L maintains consistently high true-positive counts for several structure-related and pathology-related categories, including Apical Periodontitis, Missing Tooth, Implant, Porcelain Crown, and Ceramic Bridge. It also shows a more balanced error pattern, with relatively fewer false negatives on these classes than DINOv2-L, Swin-B, ConvNeXt-B, and EffNet-B. In contrast, the competing methods exhibit more obvious confusion on difficult categories: DINOv2-L yields higher false negatives for Apical Periodontitis, Swin-B and EffNet-B show weaker discrimination on Decay and Missing Tooth, and ConvNeXt-B produces more false positives for some structure-related classes. Although Decay remains challenging for all methods, DINOv3-L still preserves a comparatively favorable trade-off between sensitivity and specificity. 

In summary, DINOv3 remains consistently strong across dental classification tasks, with especially clear gains in multi-label recognition and precision-critical settings. Although it is not the best model on every individual dataset, it provides the most balanced classification performance across panoramic and intraoral benchmarks without requiring dataset-specific backbone selection.

\begin{TabTwo}[t]

\caption{Performance comparison of supervised and self-supervised methods on classification tasks. 
}

\label{tab:cls_3settings_bigtable_50ep}

\begin{adjustbox}{max width=\linewidth}
\begin{tabular}{l cccc cccc cccc}
\toprule
\textbf{Backbone} &
\multicolumn{4}{c}{\textbf{MTMD (Single-label)}} &
\multicolumn{4}{c}{\textbf{OralXrays-9 (Multi-label)}} &
\multicolumn{4}{c}{\textbf{ADCD (Multi-label)}} \\
\cmidrule(lr){2-5}\cmidrule(lr){6-9}\cmidrule(lr){10-13}
&
\textbf{Acc}  & \textbf{Prec.}  & \textbf{Rec.}  & \textbf{F1}  &
\textbf{mAP}  & \textbf{Prec.}  & \textbf{Rec.}  & \textbf{F1}  &
\textbf{mAP}  & \textbf{Prec.}  & \textbf{Rec.}  & \textbf{F1}  \\
\midrule

\multicolumn{13}{c}{\textbf{Supervised methods}} \\
\midrule

\bk{ResNet-50}{he2016resnet}        & 90.91 & 79.72 & 82.62 & 80.97 & 94.98 & 77.36 & 95.83 & 84.60 & 68.71 & 53.12 & 76.92 & 61.85 \\
\bk{Swin-B}{liu2021swin}           & 90.91 & 78.80 & \textbf{85.71} & 81.73 & 94.99 & 75.93 & 97.25 & 84.14 & 68.46 & 51.81 & 78.33 & 61.15 \\
\bk{SwinV2-B}{liu2022swinv2}         & 84.47 & 77.54 & 59.93 & 57.83 & 91.99 & 69.32 & 97.60 & 79.52 & 64.42 & 50.39 & 79.65 & 59.39 \\
\bk{ConvNeXt-B}{liu2022convnext}       & 89.02 & 76.03 & 72.86 & 74.22 & 90.68 & 70.75 & \textbf{97.74} & 80.30 & 67.05 & 49.91 & 79.56 & 59.50 \\
\bk{EfficientNet-B4}{tan2019efficientnet}  & 76.14 & 36.49 & 42.04 & 39.06 & 91.97 & 71.03 & 96.59 & 80.43 & 66.56 & 47.39 & 77.87 & 58.02 \\
\bk{DeiT-B}{touvron2021deit}           & 79.55 & 63.52 & 48.19 & 46.34 & 84.81 & 65.88 & 95.46 & 75.88 & 63.53 & 43.07 & \textbf{84.54} & 55.18 \\
\bk{DeiT3-B}{touvron2022deit3}          & 79.17 & 44.85 & 44.10 & 43.25 & 86.77 & 70.73 & 93.27 & 79.48 & 67.58 & 49.68 & 84.34 & 59.55 \\
\bk{ViT-B}{dosovitskiy2021vit}            & 77.65 & 55.88 & 46.91 & 48.74 & 84.13 & 66.08 & 94.00 & 75.92 & 64.42 & 37.93 & 79.48 & 48.85 \\
\bk{RegNetX-8GF}{radosavovic2020regnet}      & 82.95 & 61.94 & 53.21 & 54.55 & 94.09 & 75.34 & 95.80 & 83.68 & 68.72 & 51.90 & 77.79 & 61.44 \\

\midrule
\multicolumn{13}{c}{\textbf{Self-supervised methods}} \\
\midrule

\bk{BEiT-B}{bao2021beit}          & 82.20 & 52.07 & 54.00 & 52.45 & 93.72 & \textbf{79.12} & 94.05 & 85.07 & 65.77 & 51.35 & 76.93 & 60.63 \\
\bk{DINOv2-S}{oquab2023dinov2}        & \textbf{92.05} & 87.81 & 79.24 & 82.70 & 94.97 & 75.38 & 97.52 & 83.68 & 71.21 & 55.49 & 75.43 & 62.72 \\
\bk{DINOv2-B}{oquab2023dinov2}         & 87.50 & 80.82 & 69.84 & 71.79 & 95.27 & 77.06 & 96.86 & 84.64 & 70.99 & 54.39 & 77.32 & 62.48 \\
\bk{DINOv2-L}{oquab2023dinov2}        & 85.98 & 75.51 & 59.12 & 62.56 & \textbf{95.66} & 78.87 & 97.34 & 85.85 & 72.47 & \textbf{57.42} & 74.17 & 63.65 \\
\bk{DINOv3-S}{dinov3}        & 88.26 & 80.37 & 77.68 & 77.55 & 95.08 & 78.86 & 96.44 & \textbf{85.95} & \textbf{73.06} & 55.05 & 79.83 & 63.69 \\
\bk{DINOv3-B}{dinov3}        & 89.02 & 88.05 & 69.62 & 73.05 & 95.49 & 77.59 & 96.62 & 85.12 & 70.29 & 54.85 & 84.35 & \textbf{64.43} \\
\bk{DINOv3-L}{dinov3}         & 91.67 & \textbf{88.54} & 84.30 & \textbf{85.44} & 95.47 & 77.32 & 97.48 & 85.03 & 72.54 & 57.07 & 74.86 & 64.16 \\
\bottomrule
\end{tabular}
\end{adjustbox}
\end{TabTwo}

\begin{table}[t]
\centering
\scriptsize
\setlength{\tabcolsep}{2.4pt}
\renewcommand{\arraystretch}{1.05}
\caption{
Performance comparison of supervised and self-supervised methods on intraoral multi-label classification tasks. 
}

\label{tab:ioscan_multilabel_bigtable_50ep}
\resizebox{\columnwidth}{!}{%
\begin{tabular}{l cccc cccc}
\toprule
\textbf{Backbone} &
\multicolumn{4}{c}{\textbf{DentalAI (Multi-label)}} &
\multicolumn{4}{c}{\textbf{AlphaDent (Multi-label)}} \\
\cmidrule(lr){2-5}\cmidrule(lr){6-9}
&
\textbf{mAP}  & \textbf{Prec.}  & \textbf{Rec.}  & \textbf{F1}  &
\textbf{mAP}  & \textbf{Prec.}  & \textbf{Rec.}  & \textbf{F1}  \\
\midrule

\multicolumn{9}{c}{\textbf{Supervised methods}} \\
\midrule

\bk{ResNet-50}{he2016resnet}        & 79.82 & 69.93 & 78.89 & 72.60 & 64.88 & 48.32 & 71.52 & 56.84 \\
\bk{Swin-B}{liu2021swin}           & 81.11 & 67.42 & 83.38 & 74.10 & 64.09 & 43.73 & 71.58 & 52.91 \\
\bk{SwinV2-B}{liu2022swinv2}         & 78.78 & 67.49 & 83.24 & 73.83 & 58.09 & 38.22 & 71.62 & 47.23 \\
\bk{ConvNeXt-B}{liu2022convnext}       & 80.97 & 64.92 & 88.03 & 73.17 & 58.96 & 43.63 & 69.75 & 52.61 \\
\bk{EfficientNet-B4}{tan2019efficientnet}  & 80.25 & 67.15 & 84.06 & 74.13 & 58.87 & 45.96 & 72.02 & 55.22 \\
\bk{DeiT-B}{touvron2021deit}           & 75.04 & 66.58 & 80.93 & 72.70 & 56.91 & 41.02 & 73.43 & 50.37 \\
\bk{DeiT3-B}{touvron2022deit3}          & 77.33 & 66.15 & 84.73 & 73.20 & 57.64 & 39.17 & 73.01 & 49.55 \\
\bk{ViT-B}{dosovitskiy2021vit}            & 72.00 & 64.60 & 76.88 & 69.28 & 58.86 & 48.77 & 52.63 & 50.16 \\
\bk{RegNetX-8GF}{radosavovic2020regnet}      & 78.91 & 67.31 & 86.24 & 74.67 & 63.44 & 47.14 & 68.49 & 54.69 \\

\midrule
\multicolumn{9}{c}{\textbf{Self-supervised methods}} \\
\midrule

\bk{BEiT-B}{bao2021beit}           & 80.25 & 67.12 & 85.30 & 74.34 & 56.58 & 39.72 & 60.10 & 47.02 \\
\bk{DINOv2-S}{oquab2023dinov2}         & 84.34 & 72.22 & 84.93 & 75.70 & 67.31 & 53.23 & 69.62 & 59.69 \\
\bk{DINOv2-B}{oquab2023dinov2}         & \textbf{88.14} & 72.10 & \textbf{91.97} & 80.12 & 71.40 & 59.19 & 63.03 & 60.19 \\
\bk{DINOv2-L}{oquab2023dinov2}         & 87.35 & 75.29 & 86.84 & 80.36 & \textbf{72.15} & \textbf{59.41} & \textbf{75.29} & \textbf{64.84} \\
\bk{DINOv3-S}{dinov3}         & 82.52 & 72.16 & 83.10 & 76.99 & 63.18 & 45.48 & 70.54 & 54.21 \\
\bk{DINOv3-B}{dinov3}         & 84.49 & 71.15 & 86.78 & 77.68 & 69.65 & 54.51 & 65.59 & 59.18 \\
\bk{DINOv3-L}{dinov3}         & 86.81 & \textbf{79.32} & 87.38 & \textbf{83.04} & 66.16 & 49.21 & 70.23 & 56.73 \\
\bottomrule
\end{tabular}
}
\end{table}

\subsection{Model Size and Input Resolution}\label{sec:exp_scale_res}

\subsubsection{Detection and Instance Segmentation}\label{sec:exp_det_res}

We evaluate DINOv3 at different model scales and input resolutions.
Results on panoramic radiographs and intraoral photographs are reported in Table~\ref{tab:merged_frozen_square_scales_tight_noscale} and Table~\ref{tab:dentalai_alphadent_multires_frozen_dinov3}, respectively.

Across  panoramic datasets (DENTEX, OralXrays-9, CariesXrays), performance generally improves with input resolution across panoramic datasets, although the trend is not strictly monotonic for every backbone and metric. DINOv3-L, the largest model, achieves the strongest results among the tested DINOv3 variants at the highest resolution (1280×1280). For example, on OralXrays-9, DINOv3-L at 1280×1280 attains a box AP ($AP^b$) and mask AP ($AP^m$) of 62.8\%. The performance gain from higher resolution is most dramatic on the challenging \dataset{CariesXrays} dataset: DINOv3-S's $AP^b$ improves from 10.8\% at 512×512 to 38.2\% at 1280×1280, a relative increase exceeding 250

The trend on intraoral photograph datasets (DentalAI, AlphaDent) is more nuanced. While increased resolution generally yields better results, the gains are less pronounced than with panoramic X-rays. For certain backbones, performance even plateaus or declines slightly at the maximum resolution (1280×1280), as seen with DINOv3-S on both datasets and DINOv3-B on AlphaDent. DINOv3-L performs best on \dataset{DentalAI} at 1280×1280 $AP^b$=32.2\%, $AP^m$=32.9\%). On \dataset{AlphaDent}, however, DINOv3-L achieves optimal results at 1024×1024 across most metrics. This suggests a potential trade-off between leveraging fine details and maintaining an effective receptive field for this specific dataset, indicating that the optimal resolution for intraoral photos may be task- and dataset-dependent. Unlike X-rays, these images rely more on interpreting complex textures and color variations under varied lighting.

The analysis reveals that panoramic radiograph performance shows a more monotonic and significant benefit from higher resolutions, highlighting the paramount importance of pixel-level detail for identifying subtle anatomical and pathological structures in grayscale X-rays. In contrast, segmenting intraoral photographs, while still benefiting from increased resolution, depends more heavily on the model's ability to understand complex visual appearances, with diminishing returns at very high resolutions for some tasks.
Therefore, for panoramic radiograph analysis, we recommend using the largest feasible DINOv3 backbone (DINOv3-L) at the highest practical resolution to maximize performance. For intraoral photograph analysis, a resolution of 1024×1024 often provides an excellent balance between accuracy and computational efficiency, though the optimal setting should be validated on the target dataset and task.

\begin{figure*}[t]
    \centering
    \includegraphics[width=0.75\linewidth]{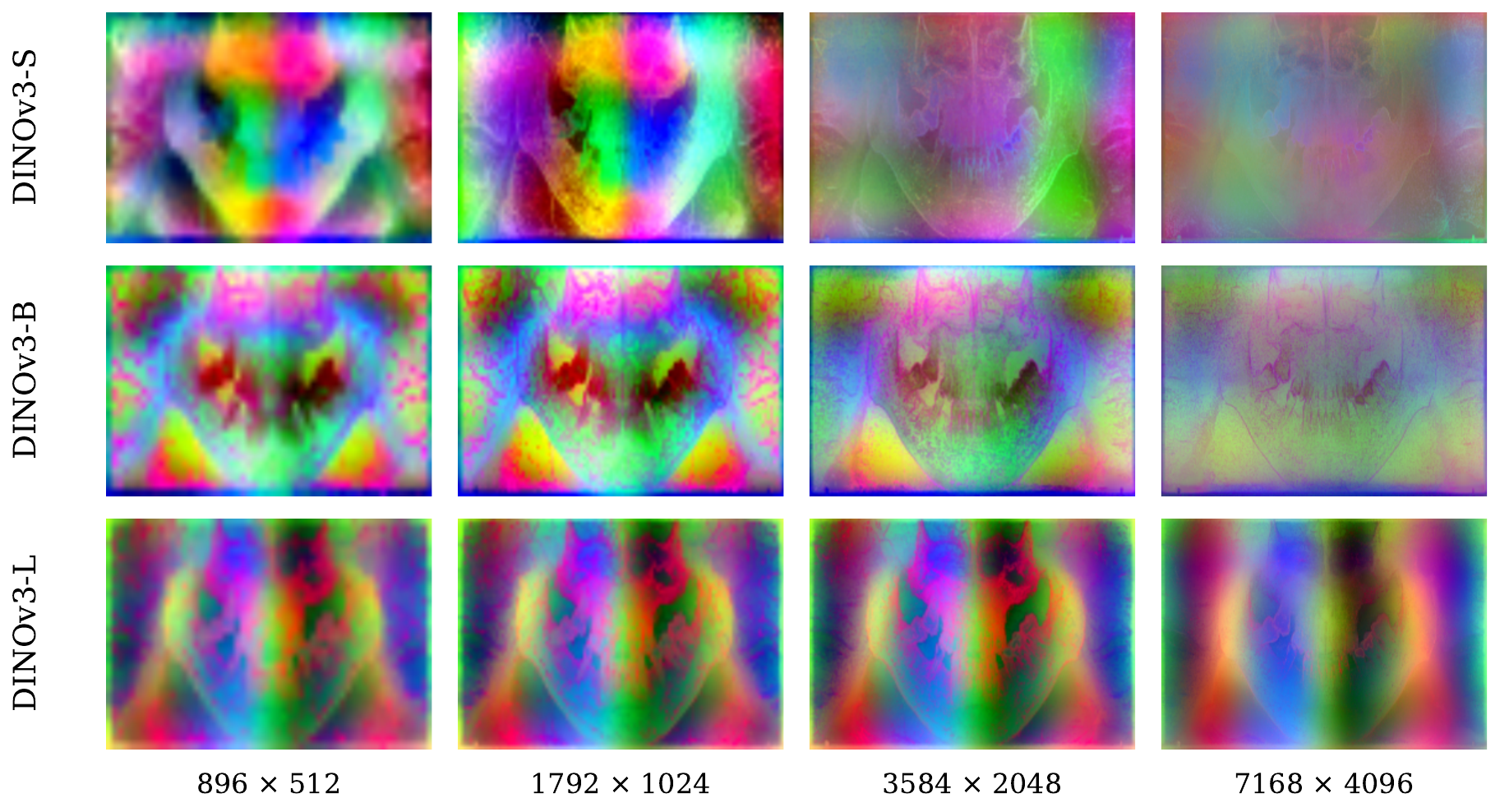}
    \caption{
    Visualization of the different resolution sensitivity of different DINOv3 sizes on panoramic radiographs.
    }
    \label{fig:fig17_vis_pano}
\end{figure*}

\begin{figure*}[t]
    \centering
    \includegraphics[width=0.75\linewidth]{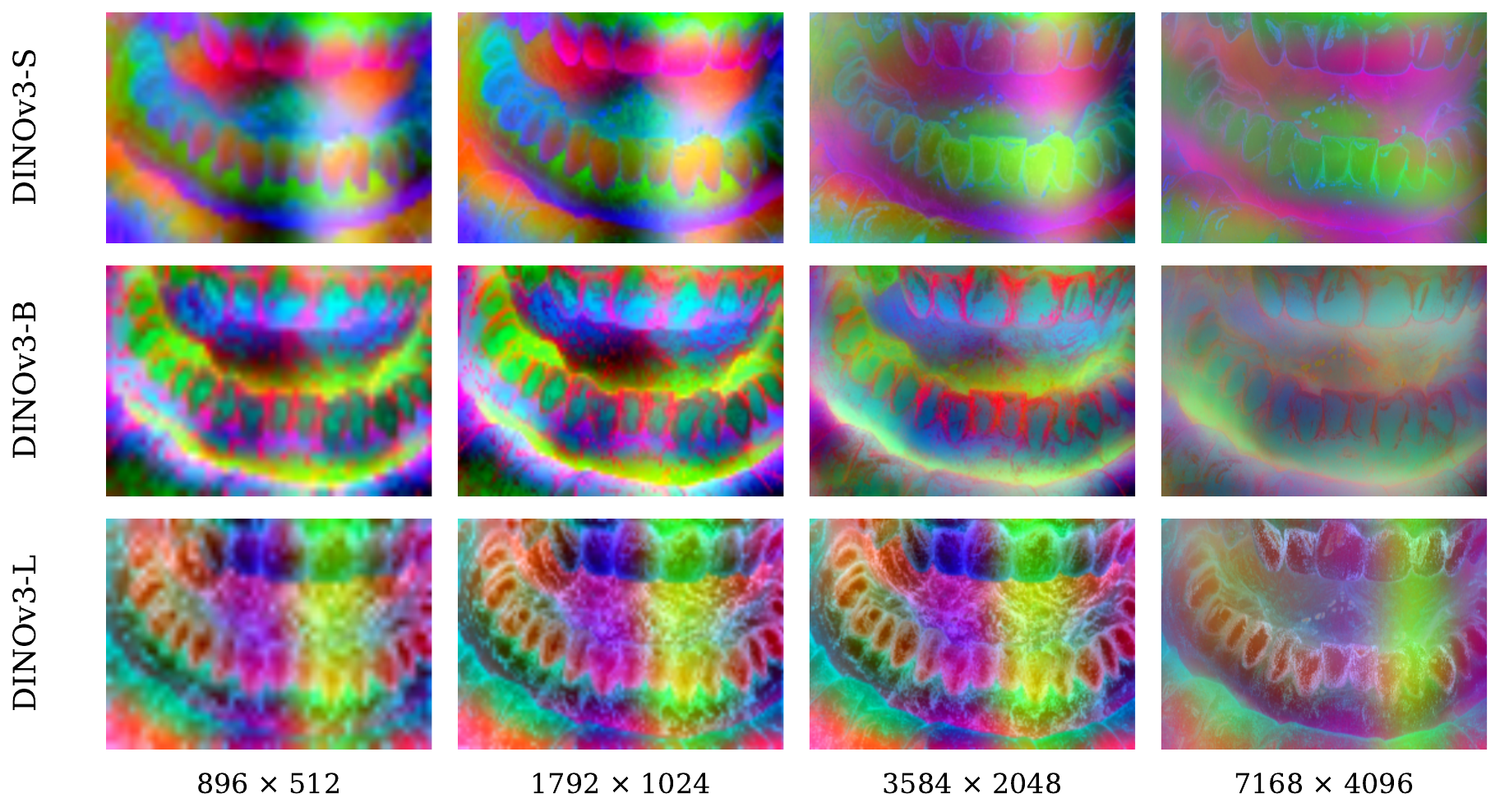}
    \caption{
    Visualization of the different resolution sensitivity of different DINOv3 sizes on intraoral photographs.
    }
    \label{fig:fig17_vis_intraoral}
\end{figure*}

\begin{TabTwo}[t]

\caption{ 
Performance comparison of different DINOv3 models and input resolutions on panoramic X-ray detection and instance segmentation datasets.
}

\label{tab:merged_frozen_square_scales_tight_noscale}
\begin{adjustbox}{max width=\linewidth}
\begin{tabular}{l l cccc cccc cccc}
\toprule
\textbf{Backbone} & \textbf{Input} &
\multicolumn{4}{c}{\textbf{DENTEX}} &
\multicolumn{4}{c}{\textbf{OralXrays-9}} &
\multicolumn{4}{c}{\textbf{CariesXrays}} \\
\cmidrule(lr){3-6}\cmidrule(lr){7-10}\cmidrule(lr){11-14}
& &
$AP^{b}$ & $AP^{b}_{50}$ & $AP^{m}$ & $AP^{m}_{50}$ &
$AP^{b}$ & $AP^{b}_{50}$ & $AP^{m}$ & $AP^{m}_{50}$ &
$AP^{b}$ & $AP^{b}_{50}$ & $AP^{m}$ & $AP^{m}_{50}$ \\
\midrule

\multirow{4}{*}{\bk{DINOv3-S}{dinov3}}  & $512 \times 512$   & 23.20 & 42.30 & 22.20 & 42.50 & 50.40 & 79.50 & 50.90 & 80.20 & 10.80 & 32.30 &  5.10 & 15.50 \\
         & $768 \times 768$   & 25.70 & 44.80 & 26.50 & 44.80 & 58.40 & 87.40 & 58.70 & 87.90 & 26.50 & 60.50 & 12.10 & 28.70 \\
         & $1024 \times 1024$  & 27.00 & 45.90 & 28.10 & 45.90 & 60.80 & 88.60 & 60.70 & 88.80 & 34.60 & 71.80 & 16.30 & 34.50 \\
         & $1280 \times 1280$  & 27.50 & 46.40 & 29.30 & 46.40 & 61.30 & 88.80 & 61.40 & 89.00 & 38.20 & 75.70 & 18.10 & 36.90 \\
\addlinespace[2pt]

\multirow{4}{*}{\bk{DINOv3-B}{dinov3}} & $512 \times 512$   & 22.80 & 43.10 & 21.90 & 41.50 & 51.10 & 80.60 & 51.40 & 81.30 &  8.00 & 24.90 &  3.80 & 11.90 \\
         & $768 \times 768$   & 24.70 & 44.90 & 25.90 & 44.80 & 58.90 & 87.60 & 59.30 & 88.20 & 27.60 & 62.20 & 12.20 & 29.30 \\
         & $1024 \times 1024$  & 27.20 & 46.80 & 28.50 & 46.50 & 61.10 & 88.90 & 61.40 & 89.20 & 35.10 & 72.90 & 16.80 & 35.80 \\
         & $1280 \times 1280$  & 27.40 & 48.00 & 28.80 & 46.80 & 60.90 & 88.80 & 61.20 & 89.10 & 38.80 & 76.80 & 18.10 & 37.60 \\
\addlinespace[2pt]

\multirow{4}{*}{\bk{DINOv3-L}{dinov3}} & $512 \times 512$   & 24.50 & 44.50 & 23.80 & 43.80 & 52.20 & 80.80 & 52.70 & 81.50 & 13.30 & 37.40 &  6.00 & 17.60 \\
         & $768 \times 768$   & 28.80 & 48.00 & 29.60 & 48.10 & 60.20 & 88.30 & 60.20 & 88.70 & 29.20 & 65.00 & 13.80 & 31.70 \\
         & $1024 \times 1024$  & 30.00 & 50.20 & 31.60 & \best{51.00} & 62.10 & \best{89.50} & 62.10 & 89.80 & 37.80 & 75.40 & 17.40 & 36.30 \\
         & $1280 \times 1280$  & \best{30.30} & \best{51.10} & \best{32.20} & 50.00 & \best{62.80} & \best{89.50} & \best{62.80} & \best{89.90} & \best{41.30} & \best{79.30} & \best{19.70} & \best{39.10} \\
\bottomrule
\end{tabular}
\end{adjustbox}
\end{TabTwo}

\begin{table}[t]
\centering
\scriptsize
\setlength{\tabcolsep}{2.4pt} 
\renewcommand{\arraystretch}{1.05}

\caption{
Performance comparison of different DINOv3 models and input resolutions on intraoral detection and instance segmentation datasets.
}

\label{tab:dentalai_alphadent_multires_frozen_dinov3}
\resizebox{\columnwidth}{!}{%
\begin{tabular}{l l cccc cccc}
\toprule
\textbf{Backbone} & \textbf{Input} &
\multicolumn{4}{c}{\textbf{DentalAI}} &
\multicolumn{4}{c}{\textbf{AlphaDent}} \\
\cmidrule(lr){3-6}\cmidrule(lr){7-10}
& &
$AP^{b}$ & $AP^{b}_{50}$ & $AP^{m}$ & $AP^{m}_{50}$ &
$AP^{b}$ & $AP^{b}_{50}$ & $AP^{m}$ & $AP^{m}_{50}$ \\
\midrule

\multirow{4}{*}{\bk{DINOv3-S}{dinov3}}
& $512 \times 512$  & 25.50 & 42.10 & 25.40 & 40.60 & 17.90 & 32.80 & 17.20 & 30.50 \\
& $768 \times 768$  & 28.60 & 45.20 & 28.90 & 44.40 & 21.50 & 38.70 & 20.90 & 36.00 \\
& $1024 \times 1024$ & 29.20 & 49.30 & 30.30 & 46.60 & 23.50 & 43.80 & 23.30 & 42.40 \\
& $1280 \times 1280$ & 28.40 & 47.20 & 29.80 & 46.50 & 22.10 & 39.20 & 22.20 & 37.60 \\
\addlinespace[2pt]

\multirow{4}{*}{\bk{DINOv3-B}{dinov3}}
& $512 \times 512$  & 28.20 & 46.10 & 27.90 & 44.30 & 19.90 & 35.50 & 18.70 & 30.90 \\
& $768 \times 768$  & 29.20 & 48.10 & 29.50 & 45.00 & 22.20 & 39.40 & 21.40 & 38.10 \\
& $1024 \times 1024$ & 28.70 & 47.30 & 30.10 & 46.30 & 23.70 & 43.30 & 22.40 & 40.00 \\
& $1280 \times 1280$ & 29.60 & 48.60 & 30.20 & 47.80 & 23.50 & 41.60 & 22.80 & 40.20 \\
\addlinespace[2pt]

\multirow{4}{*}{\bk{DINOv3-L}{dinov3}}
& $512 \times 512$  & 28.50 & 45.70 & 27.10 & 42.80 & 21.00 & 37.20 & 19.90 & 33.10 \\
& $768 \times 768$  & 30.20 & 48.70 & 29.90 & 46.50 & 22.70 & 40.90 & 22.00 & 38.20 \\
& $1024 \times 1024$ & 31.60 & 51.10 & 32.30 & 48.70 & 25.40 & \best{44.60} & 24.20 & \best{42.90} \\
& $1280 \times 1280$ & \best{32.20} & \best{52.00} & \best{32.90} & \best{50.20} & 23.90 & 43.30 & 23.80 & 41.80 \\
\bottomrule
\end{tabular}
}
\end{table}


\subsubsection{Semantic Segmentation}\label{sec:exp_seg_scale}




In Table~\ref{tab:peri_impacted_frozen_square_scales}, we examine the influence of DINOv3 model size and input resolution on semantic segmentation performance for two dental tasks.

A consistent, non-monotonic trend is observed across all model sizes (S, B, L) and datasets: performance improves significantly as resolution increases from $512\times512$ to $1024\times1024$, but then slightly \textit{declines} when moving to $1280\times1280$. The resolution of $1024\times1024$ emerges as the clear optimum for semantic segmentation on these dental tasks. For example, on the more challenging \dataset{Impacted} dataset, DINOv3-S achieves its peak mIoU of 86.66\% at $1024\times1024$, which drops to 85.30\% at $1280\times1280$.  The benefit of increasing resolution is most dramatic on the \dataset{Impacted} dataset, which is inherently more difficult as indicated by its lower overall scores. For DINOv3-S, increasing resolution from $512$ to $1024$ yields a gain of over 10 points in mIoU (76.36\% to 86.66\%). In contrast, on the \dataset{Periodontitis} dataset, where performance is already saturated at high levels (mIoU $> 90\%$ even at $512$), the absolute gains from higher resolution are smaller but still crucial for reaching near-perfect segmentation (mIoU $> 95\%$). This indicates that sufficient resolution is critical for the model to delineate fine anatomical structures or pathological regions accurately.

While larger backbones (DINOv3-L) generally achieve the best absolute performance at the optimal resolution (e.g., 87.08\% mIoU on \dataset{Impacted}), the scaling trend is remarkably consistent across all model sizes. All three variants (S, B, L) peak at $1024\times1024$ and show a slight decline at $1280\times1280$. This suggests that the observed optimal resolution is a property of the segmentation task and head architecture, rather than being specific to a particular backbone capacity. At very high input resolutions, the effective receptive field and contextual understanding per patch might become suboptimal for making precise, locally consistent pixel-level decisions, potentially leading to slightly noisier segmentation masks despite the increased input detail. This suggests that once sufficient spatial detail is provided, panoramic semantic segmentation is governed more by adequate spatial sampling of the input than by further increases in backbone capacity.

\begin{table}[t]
\centering
\scriptsize
\setlength{\tabcolsep}{2.4pt} 
\renewcommand{\arraystretch}{1.05}
    \caption{ 
    Performance comparison of different DINOv3 models and input resolutions on the semantic segmentation task.
    }

	\label{tab:peri_impacted_frozen_square_scales}
\resizebox{\columnwidth}{!}{%
	\begin{tabular}{l l cccc cccc}
		\toprule
		\textbf{Backbone} & \textbf{Input} &
		\multicolumn{4}{c}{\textbf{Periodontitis}} &
		\multicolumn{4}{c}{\textbf{Impacted}} \\
		\cmidrule(lr){3-6}\cmidrule(lr){7-10}
		& &
		\textbf{mIoU}  & \textbf{mDice}  & \textbf{Prec.}  & \textbf{Rec.}  &
		\textbf{mIoU}  & \textbf{mDice}  & \textbf{Prec.}  & \textbf{Rec.}  \\
		\midrule
		
		\multirow{4}{*}{\bk{DINOv3-S}{dinov3}}
		& $512 \times 512$   & 90.84 & 95.05 & 94.30 & 95.83 & 76.36 & 84.63 & 83.78 & 85.53 \\
		& $768 \times 768$   & 91.10 & 95.20 & 94.80 & 95.61 & 77.26 & 85.38 & 84.73 & 86.06 \\
		& $1024 \times 1024$  & 95.81 & 97.83 & 98.03 & 97.63 & 86.66 & 92.33 & 93.01 & 91.67 \\
		& $1280 \times 1280$  & 95.27 & 97.54 & 97.72 & 97.36 & 85.30 & 91.42 & 92.85 & 90.08 \\
		\addlinespace[2pt]
		
		\multirow{4}{*}{\bk{DINOv3-B}{dinov3}}
		& $512 \times 512$   & 90.28 & 94.72 & 93.95 & 95.53 & 75.93 & 84.26 & 84.58 & 83.95 \\
		& $768 \times 768$   & 91.50 & 95.43 & 95.51 & 95.36 & 77.58 & 85.65 & 85.69 & 85.60 \\
		& $1024 \times 1024$  & \best{95.88} & \best{97.87} & 97.98 & \best{97.76} & 86.52 & 92.24 & 92.73 & 91.76 \\
		& $1280 \times 1280$  & 95.53 & 97.68 & 98.05 & 97.32 & 85.20 & 91.35 & 92.53 & 90.23 \\
		\addlinespace[2pt]
		
		\multirow{4}{*}{\bk{DINOv3-L}{dinov3}}
		& $512 \times 512$   & 90.71 & 94.97 & 94.63 & 95.32 & 76.05 & 84.37 & 84.28 & 84.46 \\
		& $768 \times 768$   & 91.25 & 95.29 & 96.25 & 94.37 & 77.55 & 85.62 & 85.84 & 85.40 \\
		& $1024 \times 1024$  & 95.87 & 97.86 & \best{98.16} & 97.57 & \best{87.08} & \best{92.60} & 93.29 & \best{91.94} \\
		& $1280 \times 1280$  & 95.49 & 97.66 & 97.61 & 97.70 & 85.95 & 91.86 & \best{93.48} & 90.35 \\
		\bottomrule
	\end{tabular}
}
\end{table}


\subsubsection{Classification}\label{sec:exp_cls_scale}




In Table~\ref{tab:mtmd_oralx_adcd_multires_4dp}, we evaluate the influence of different sizes of DINOv3 and input resolutions under the task of classification.

In contrast to dense prediction, classification exhibits a much weaker and less monotonic dependence on input resolution.
Increasing the input size can bring improvements, but the best performance does not consistently occur at the highest resolution, and the trend varies substantially across datasets and metrics.
This suggests that once global semantic content is preserved, further increasing spatial detail brings limited and often unstable gains for dental classification.

For the single-label dataset \dataset{MTMD}, resolution scaling is clearly non-monotonic across all model sizes.
For example, DINOv3-B achieves the best Accuracy of 80.68\% at $768\times768$, rather than at $1024\times1024$, while DINOv3-L reaches only 79.17\% at both $768\times768$ and $1024\times1024$.
A similar pattern is observed for macro F1, where the overall best result is 42.96\% for DINOv3-B at $224\times224$.
These results indicate that for relatively coarse-grained single-label recognition, higher resolution is not the dominant factor once the main anatomical structure is already visible.

For the multi-label panoramic datasets \dataset{OralXrays-9} and \dataset{ADCD}, the resolution effect is somewhat stronger, but the improvement is still not consistently monotonic.
On \dataset{OralXrays-9}, moderate resolutions generally perform best: DINOv3-L reaches its highest mAP of 80.52\% at $768\times768$, while its best F1 of 70.13\% is obtained at $512\times512$.
Similarly, DINOv3-B achieves its best mAP of 79.79\% at $512\times512$ and best F1 of 69.56\% at $768\times768$.
Performance at $1024\times1024$ is not better, and in several cases is slightly worse, indicating that the benefit of higher resolution saturates at moderate input sizes.

On the more challenging \dataset{ADCD} dataset, the trend is even more dataset-dependent.
For DINOv3-S, both mAP and F1 peak at $768\times768$ (64.38\% and 55.54\%, respectively), then decline at $1024\times1024$.
For DINOv3-B, the best mAP is also obtained at $768\times768$ (64.50\%), whereas Recall is highest at $512\times512$ (80.08\%).
More notably, DINOv3-L achieves its best mAP of 70.38\% already at the lowest resolution, $224\times224$, and its F1 varies only marginally across all tested resolutions.
This suggests that on ADCD, backbone scale can already provide sufficiently strong global representations, such that increasing input resolution does not translate into consistent gains.

Overall, classification behaves differently from detection and segmentation.
Whereas dense prediction benefits strongly from recovering fine spatial detail, classification depends more on global semantic recognition and therefore shows much weaker sensitivity to aggressive resolution scaling.
In practice, moderate resolutions such as $512\times512$ or $768\times768$ are usually sufficient, while moving to the largest tested resolution rarely yields consistent improvements.
Model scale also provides only limited and dataset-dependent gains, with DINOv3-L not uniformly outperforming DINOv3-B across all settings.

\begin{TabTwo}[t]

\caption{ 
Performance comparison of different DINOv3 models and input resolutions on the classification task.
}

		\label{tab:mtmd_oralx_adcd_multires_4dp}
\begin{adjustbox}{max width=\linewidth}
\begin{tabular}{l l cccc cccc cccc}
\toprule
\textbf{Backbone} & \textbf{Input} &
\multicolumn{4}{c}{\textbf{MTMD}} &
\multicolumn{4}{c}{\textbf{OralXrays-9}} &
\multicolumn{4}{c}{\textbf{ADCD}} \\
\cmidrule(lr){3-6}\cmidrule(lr){7-10}\cmidrule(lr){11-14}
& &
\textbf{Acc} & \textbf{Prec.} & \textbf{Rec.} & \textbf{F1} &
\textbf{mAP} & \textbf{Prec.} & \textbf{Rec.} & \textbf{F1} &
\textbf{mAP} & \textbf{Prec.} & \textbf{Rec.} & \textbf{F1} \\
\midrule

\multirow{4}{*}{\bk{DINOv3-S}{dinov3}}
& $224 \times 224$  & 76.14 & 36.63 & 40.79 & 38.54  & 76.34 & 54.54 & 95.79 & 67.86  & 62.00 & 41.94 & 74.19 & 51.73 \\
& $512 \times 512$  & 78.03 & 37.89 & 42.40 & 39.98  & 78.99 & 56.16 & 96.08 & 68.96  & 64.84 & 41.52 & 78.50 & 52.45 \\
& $768 \times 768$  & 78.03 & 37.98 & 43.02 & 40.34  & 78.51 & 56.28 & 95.83 & 68.90  & 64.38 & \best{46.59} & 78.56 & \best{55.54} \\
& $1024 \times 1024$ & 78.03 & 37.74 & 43.02 & 40.21  & 77.50 & 55.09 & 96.02 & 68.06  & 62.85 & 43.81 & 78.00 & 53.59 \\
\addlinespace[2pt]

\multirow{4}{*}{\bk{DINOv3-B}{dinov3}}
& $224 \times 224$  & 78.03 & \best{42.45} & \best{44.67} & \best{42.96}  & 78.16 & 54.88 & 96.17 & 68.40  & 63.51 & 41.73 & 77.97 & 53.54 \\
& $512 \times 512$  & 79.17 & 38.58 & 44.11 & 41.16  & 79.79 & 55.93 & 95.77 & 69.18  & 64.05 & 42.13 & \best{80.08} & 53.73 \\
& $768 \times 768$  & \best{80.68} & 39.89 & 44.31 & 41.95  & 79.47 & 56.91 & 94.90 & 69.56  & 64.50 & 41.34 & 77.76 & 53.33 \\
& $1024 \times 1024$ & 80.30 & 39.22 & 44.57 & 41.72  & 78.93 & 56.00 & 95.45 & 68.96  & 64.24 & 44.32 & 72.55 & 52.98 \\
\addlinespace[2pt]

\multirow{4}{*}{\bk{DINOv3-L}{dinov3}}
& $224 \times 224$  & 76.52 & 37.63 & 41.16 & 39.17  & 78.79 & 55.91 & \best{96.38} & 69.08  & \best{70.38} & 43.12 & 79.30 & 54.42 \\
& $512 \times 512$  & 78.03 & 37.62 & 43.85 & 40.43  & 80.22 & \best{57.50} & 96.12 & \best{70.13}  & 62.87 & 45.07 & 77.77 & 55.08 \\
& $768 \times 768$  & 79.17 & 38.92 & 44.11 & 41.35  & \best{80.52} & 57.34 & 96.18 & 69.96  & 63.10 & 43.82 & 77.53 & 54.45 \\
& $1024 \times 1024$ & 79.17 & 38.36 & 44.11 & 41.03  & 79.49 & 56.43 & 96.11 & 68.95  & 63.82 & 44.29 & 77.89 & 54.67 \\
\bottomrule
\end{tabular}%

\end{adjustbox}
\end{TabTwo}



Across \emph{dense prediction} tasks, input resolution is the dominant factor: performance typically improves substantially from $512 \times 512$ to around $1024 \times 1024$, where most datasets reach a practical operating point, while further increasing to $1280 \times 1280$ yields diminishing returns and can become mildly non-monotonic.
This behavior is strongly \emph{dataset- and modality-dependent}: tiny, low-contrast lesions such as those in \dataset{CariesXrays} are the most resolution-sensitive, whereas larger or better-separated targets saturate earlier.
Model scale provides an additional but secondary benefit, which becomes more visible only after the spatial bottleneck has been sufficiently relieved.
By contrast, \emph{classification} exhibits much weaker and less stable resolution dependence, suggesting that moderate inputs are usually sufficient for global dental recognition.
Taken together, these results show that the standard scaling rule of ``larger model + higher resolution'' transfers to dentistry only up to a point.
This motivates the next subsection, where we examine whether adaptation strategy---rather than further scaling alone---is the more effective lever for closing the remaining dental domain gap.

\subsection{Adaptation Strategy}\label{sec:exp_adapt}


We investigate how DINOv3 can be adapted to the dental domain in practice. To this end, we compare three representative strategies under matched computational budgets: Frozen, Full Fine-tuning, and the parameter-efficient LoRA (rank=8). Our goal is not only to compare their absolute performance, but also to identify which strategy best balances domain adaptation, model robustness, and training efficiency, given the substantial domain shift present in dental data.In the panoramic tables, we report the number of trainable parameters (TrP); for the intraoral tables, it is omitted as the values are nearly identical.

\subsubsection{Detection and Instance Segmentation}\label{sec:exp_det_tuning}

In Table~\ref{tab:maskrcnn_4datasets_frozen_ft_lora8_singlebest} and Table~\ref{tab:maskrcnn_dentai_alphadent_frozen_ft_lora8_singlebest}, we compare different adaptation strategies for DINOv3 on detection and instance segmentation tasks.

On all four panoramic datasets (\dataset{DENTEX}, \dataset{ADCD}, \dataset{CariesXrays}, \dataset{OralXrays-9}), adapting the frozen DINOv3 features is crucial for strong performance. Among the three strategies, \textbf{LoRA consistently provides competitive results, especially for larger models, while maintaining strong parameter efficiency}. For the largest backbone, DINOv3-L, LoRA yields some of the strongest results in this comparison, performing best on DENTEX, ADCD, and OralXrays-9 while remaining competitive on CariesXrays, all while tuning only 36M parameters. This is significantly fewer than the 323M parameters required for full Finetune. The improvement over the Frozen baseline is substantial; for instance, on DENTEX, DINOv3-L with LoRA ($AP^b=35.4\%$) outperforms its Frozen counterpart ($AP^b=30.4\%$) by 5.0 points. Notably, full Finetune, while effective, sometimes underperforms relative to LoRA (e.g., on OralXrays-9 for DINOv3-L) or shows diminishing returns given its high computational cost, indicating potential overfitting on smaller medical datasets.

The trends on intraoral datasets (\dataset{DentalAI} and \dataset{AlphaDent}) are more nuanced, reflecting a potentially smaller domain shift from DINOv3's natural image pre-training to photographs compared to X-rays. \textbf{LoRA and Finetune both provide significant gains over the Frozen baseline, but their relative effectiveness varies}. On \dataset{DentalAI}, both Finetune and LoRA lead to strong improvements, with Finetune having a slight edge for DINOv3-B and DINOv3-L. However, on the more challenging \dataset{AlphaDent} dataset, two key observations stand out: 1) The Frozen baseline is relatively stronger, and 2) Full Finetune can occasionally degrade performance compared to the Frozen baseline (e.g., DINOv3-L's $AP^m$ drops from 26.5\% to 25.3\%). In contrast, LoRA provides more stable and reliable improvements, achieving the best $AP^b$ on AlphaDent for all backbone sizes. This suggests that the conservative, low-rank update of LoRA is particularly beneficial for adapting to specialized photographic domains without disrupting the valuable pre-trained representations.

Therefore, for \textbf{Panoramic Radiography}, characterized by a significant domain shift (natural images to X-rays), \textbf{LoRA appears to be a strong practical choice under our benchmark setting}. It reliably extracts near-maximal performance from large backbones like DINOv3-L with minimal parameter overhead, making it highly efficient for deployment. For \textbf{Intraoral Photography}, where the domain gap may be narrower, \textbf{LoRA remains a robust and safe choice}, especially on difficult datasets. Full Finetune can be considered if computational resources are abundant and the target dataset is large enough to mitigate overfitting risks, as it may yield marginal gains on some benchmarks. The \textbf{Frozen} strategy, while highly parameter-efficient, consistently underperforms adaptive methods, confirming that some degree of task-specific adaptation is necessary for high-precision dental object detection.

In summary, LoRA emerges as a strong parameter-efficient tuning method for dental object detection. It effectively bridges the domain gap for panoramic X-rays and provides stable, efficient adaptation for intraoral photos, establishing an optimal balance between performance, parameter efficiency, and training stability.

\begin{TabTwo}[t]

\caption{Detection and instance segmentation results with Frozen / Fine-tuning / LoRA on four panoramic datasets.}

\label{tab:maskrcnn_4datasets_frozen_ft_lora8_singlebest}

\begin{adjustbox}{max width=\linewidth}
\begin{tabular}{l l >{\centering\arraybackslash}p{9mm} cccc cccc cccc cccc}
\toprule
\textbf{Backbone} & \textbf{Tuning} & \textbf{TrP} &
\multicolumn{4}{c}{\textbf{DENTEX}} &
\multicolumn{4}{c}{\textbf{ADCD}} &
\multicolumn{4}{c}{\textbf{CariesXrays}} &
\multicolumn{4}{c}{\textbf{OralXrays-9}} \\
\cmidrule(lr){4-7}\cmidrule(lr){8-11}\cmidrule(lr){12-15}\cmidrule(lr){16-19}
& & &
$AP^{b}$ & $AP^{b}_{50}$ & $AP^{m}$ & $AP^{m}_{50}$ &
$AP^{b}$ & $AP^{b}_{50}$ & $AP^{m}$ & $AP^{m}_{50}$ &
$AP^{b}$ & $AP^{b}_{50}$ & $AP^{m}$ & $AP^{m}_{50}$ &
$AP^{b}$ & $AP^{b}_{50}$ & $AP^{m}$ & $AP^{m}_{50}$ \\
\midrule

\multirow{3}{*}{\bk{DINOv3-S}{dinov3}}
& Frozen   & 20M &
26.30 & 45.50 & 27.80 & 45.20 &
16.70 & 41.50 & 17.20 & 42.30 &
38.80 & 75.90 & 18.40 & 37.00 &
61.20 & 88.90 & 61.70 & 89.10 \\
& Finetune & 41M &
30.60 & 48.80 & 31.30 & 48.60 &
18.70 & 44.50 & 18.90 & 44.90 &
40.80 & 77.00 & 18.80 & 37.80 &
66.10 & 90.30 & 66.20 & 90.40 \\
& LoRA     & 22M &
28.00 & 46.50 & 29.20 & 46.70 &
18.10 & 44.80 & 18.10 & 44.70 &
43.60 & 80.50 & 20.20 & \best{39.50} &
66.70 & 90.60 & 66.30 & 90.80 \\
\midrule

\multirow{3}{*}{\bk{DINOv3-B}{dinov3}}
& Frozen   & 20M &
27.40 & 48.60 & 30.10 & 47.80 &
17.70 & 43.70 & 18.10 & 44.30 &
38.70 & 76.30 & 18.10 & 37.30 &
61.40 & 89.00 & 61.60 & 89.20 \\
& Finetune & 100M &
31.80 & 51.40 & 32.60 & 50.80 &
19.20 & 45.10 & 19.40 & 45.20 &
41.90 & 79.00 & 19.70 & 38.80 &
65.90 & 89.80 & 65.70 & 89.80 \\
& LoRA     & 28M &
30.70 & 49.80 & 31.40 & 49.90 &
19.70 & 45.80 & 19.80 & 46.30 &
\best{43.80} & 79.90 & 20.00 & 38.80 &
67.00 & \best{90.60} & 66.70 & \best{90.90} \\
\midrule

\multirow{3}{*}{\bk{DINOv3-L}{dinov3}}
& Frozen   & 20M &
30.40 & 50.80 & 32.30 & 50.70 &
18.80 & 45.70 & 19.30 & 46.40 &
41.60 & 79.90 & 19.30 & 38.50 &
62.80 & 89.60 & 62.90 & 89.90 \\
& Finetune & 323M &
34.40 & 54.70 & 35.60 & 55.30 &
19.00 & 44.30 & 18.80 & 43.60 &
43.30 & 78.90 & \best{20.40} & 38.70 &
65.80 & 89.30 & 65.60 & 89.40 \\
& LoRA     & 36M &
\best{35.40} & \best{55.70} & \best{36.40} & \best{55.60} &
\best{20.10} & \best{46.40} & \best{19.90} & \best{46.70} &
43.50 & \best{80.50} & 20.20 & 39.10 &
\best{67.20} & 90.40 & \best{66.90} & 90.40 \\
\bottomrule
\end{tabular}%

\end{adjustbox}
\end{TabTwo}

\begin{table}[t]
\centering
\scriptsize
\setlength{\tabcolsep}{2.4pt} 
\renewcommand{\arraystretch}{1.05}

\caption{Detection and instance segmentation results with Frozen / Fine-tuning / LoRA on two intraoral datasets.}
\label{tab:maskrcnn_dentai_alphadent_frozen_ft_lora8_singlebest}

\resizebox{\columnwidth}{!}{%
\begin{tabular}{l l cccc cccc}
\toprule
\textbf{Backbone} & \textbf{Tuning} &
\multicolumn{4}{c}{\textbf{DentalAI}} &
\multicolumn{4}{c}{\textbf{AlphaDent}} \\
\cmidrule(lr){3-6}\cmidrule(lr){7-10}
& &
$AP^{b}$ & $AP^{b}_{50}$ & $AP^{m}$ & $AP^{m}_{50}$ &
$AP^{b}$ & $AP^{b}_{50}$ & $AP^{m}$ & $AP^{m}_{50}$ \\
\midrule

\multirow{3}{*}{\bk{DINOv3-S}{dinov3}}
& Frozen   & 28.40 & 46.80 & 29.60 & 46.20 & 21.50 & 40.50 & 21.60 & 38.00 \\
& Finetune & 33.90 & 53.70 & 33.60 & 51.50 & 25.00 & 45.70 & 24.40 & 43.00 \\
& LoRA     & 33.00 & 52.30 & 33.10 & 50.70 & 25.40 & 44.60 & 24.20 & 41.70 \\
\midrule

\multirow{3}{*}{\bk{DINOv3-B}{dinov3}}
& Frozen   & 28.90 & 48.50 & 30.00 & 46.90 & 21.90 & 40.60 & 22.20 & 37.90 \\
& Finetune & 35.70 & 55.30 & 34.50 & 51.90 & 25.50 & 43.80 & 24.60 & 42.60 \\
& LoRA     & 35.00 & 55.60 & 34.50 & 53.40 & 26.00 & 45.90 & 24.90 & 42.80 \\
\midrule

\multirow{3}{*}{\bk{DINOv3-L}{dinov3}}
& Frozen   & 32.90 & 53.30 & 33.20 & 52.00 & 26.70 & 48.10 & \best{26.50} & \best{45.90} \\
& Finetune & 36.70 & \best{57.80} & \best{35.80} & \best{55.70} & 26.50 & 47.00 & 25.30 & 42.80 \\
& LoRA     & \best{37.20} & 56.60 & 35.60 & 54.00 & \best{27.50} & \best{48.60} & 25.70 & 44.60 \\
\bottomrule
\end{tabular}%

}
\end{table}


\subsubsection{Semantic Segmentation}\label{sec:exp_seg_tuning}

In Table~\ref{tab:upernet_3datasets_frozen_ft_lora_r8_bestpercolumn}, we compare parameter-efficient tuning strategies for DINOv3 models of different scales on semantic segmentation.

Across all model sizes and datasets, the \textbf{LoRA} strategy consistently delivers performance on par with, and in several cases surpassing, the full \textbf{Finetune}, while utilizing orders of magnitude fewer trainable parameters (TrP). For the largest model, DINOv3-L, LoRA (9M TrP) matches the mIoU of Finetune (309M TrP) on both \dataset{Impacted} (89.00\% vs. 89.04\%) and \dataset{Periodontitis} (96.36\% vs. 96.32\%), and comes within 0.2 points on \dataset{Caries} (78.47\% vs. 78.66\%). This pattern also holds for DINOv3-S and DINOv3-B, suggesting that LoRA can offer a highly favorable parameter-efficiency profile while maintaining competitive segmentation accuracy. The Frozen baseline, while extremely lightweight, shows a clear and consistent performance gap, particularly on the more challenging \dataset{Caries} dataset, confirming the necessity of task-specific adaptation even for strong pre-trained features.

The absolute benefit of adaptation (via Finetune or LoRA) over the Frozen baseline is heavily dependent on task difficulty. Gains are marginal on the near-saturated \dataset{Periodontitis} task (e.g., mIoU improvements $<0.5$ points for DINOv3-L) but substantial on \dataset{Caries} (e.g., $>5$ points gain in mIoU for all models). Notably, LoRA often improves the \textbf{precision-recall balance} compared to Finetune. On \dataset{Impacted}, DINOv3-L with LoRA achieves a significantly higher Precision (94.50\%) than with Finetune (93.09\%), while maintaining strong Recall (93.19\%). This suggests that LoRA's constrained, low-rank updates provide a regularizing effect, potentially leading to more confident and precise pixel-level predictions compared to the less constrained full Finetune.

The results suggest that \textbf{LoRA is a strong practical choice for semantic segmentation in dental imaging under our benchmark setting}. It effectively bridges the performance gap between frozen features and full finetuning, achieving state-of-the-art results (e.g., best mIoU for DINOv3-S on \dataset{Impacted} and best Precision for DINOv3-L on \dataset{Impacted}) at a fraction of the adaptive parameters. This makes it ideal for scenarios with limited computational resources or when deploying multiple specialized models, as the massive backbone weights remain shared and static, with only minimal, task-specific LoRA adapters being swapped.

\begin{TabTwo}[t]

\caption{Semantic segmentation results with Frozen / Fine-tuning / LoRA on three panoramic datasets.
}

\label{tab:upernet_3datasets_frozen_ft_lora_r8_bestpercolumn}

\begin{adjustbox}{max width=\linewidth}
	\begin{tabular}{l l >{\centering\arraybackslash}p{8mm} cccc cccc cccc cccc}
\toprule
\textbf{Backbone} & \textbf{Tuning} & \textbf{TrP} &
\multicolumn{4}{c}{\textbf{Impacted}} &
\multicolumn{4}{c}{\textbf{Periodontitis}} &
\multicolumn{4}{c}{\textbf{Caries}} \\
\cmidrule(lr){4-7}\cmidrule(lr){8-11}\cmidrule(lr){12-15}
& & &
\textbf{mIoU}  & \textbf{mDice}  & \textbf{Prec.}  & \textbf{Rec.}  &
\textbf{mIoU}  & \textbf{mDice}  & \textbf{Prec.}  & \textbf{Rec.}  &
\textbf{mIoU}  & \textbf{mDice}  & \textbf{Prec.}  & \textbf{Rec.}  \\
\midrule

\multirow{3}{*}{\bk{DINOv3-S}{dinov3}}
& Frozen   & 5M   &
86.66 & 92.33 & 93.01 & 91.67 &
95.81 & 97.83 & 98.03 & 97.63 &
69.69 & 78.45 & 88.02 & 72.78 \\
& Finetune & 27M  &
87.90 & 93.14 & 93.43 & 92.85 &
96.02 & 97.94 & 97.93 & 97.96 &
76.56 & 84.80 & 90.22 & 80.69 \\
& LoRA     & 6M   &
88.18 & 93.32 & 93.91 & 92.74 &
96.00 & 97.93 & 98.08 & 97.79 &
75.92 & 84.25 & 89.86 & 80.05 \\
\midrule

\multirow{3}{*}{\bk{DINOv3-B}{dinov3}}
& Frozen   & 6M   &
86.52 & 92.24 & 92.73 & 91.76 &
95.88 & 97.87 & 97.98 & 97.76 &
71.95 & 80.67 & 87.07 & 76.19 \\
& Finetune & 92M  &
88.68 & 93.63 & 92.64 & \best{94.67} &
96.22 & 98.05 & 98.08 & 98.02 &
77.37 & 85.47 & 89.26 & 82.36 \\
& LoRA     & 7M   &
88.54 & 93.54 & 93.26 & 93.83 &
96.14 & 98.00 & 98.17 & 97.84 &
77.51 & 85.59 & \best{90.56} & 81.72 \\
\midrule

\multirow{3}{*}{\bk{DINOv3-L}{dinov3}}
& Frozen   & 6M   &
87.08 & 92.60 & 93.29 & 91.94 &
95.87 & 97.86 & 98.16 & 97.57 &
73.08 & 81.73 & 87.39 & 77.59 \\
& Finetune & 309M &
\best{89.04} & \best{93.86} & 93.09 & 94.66 &
96.32 & 98.10 & \best{98.22} & 97.99 &
\best{78.66} & \best{86.52} & 89.97 & \best{83.62} \\
& LoRA     & 9M   &
89.00 & 93.84 & \best{94.50} & 93.19 &
\best{96.36} & \best{98.12} & 98.20 & \best{98.05} &
78.47 & 86.36 & 90.23 & 83.19 \\
\bottomrule
\end{tabular}%

\end{adjustbox}
\end{TabTwo}

\subsubsection{Classification}\label{sec:exp_cls_tuning}

In Table~\ref{tab:cls_3datasets_frozen_finetune_lora_n8} and Table~\ref{tab:ml_cls_2datasets_frozen_finetune_lora_n8}, we compare the performance of different tuning strategies for DINOv3 models on classification tasks.

Across all datasets and model sizes, a stark performance gap exists between the \textbf{Frozen} baseline and adaptive methods (\textbf{Finetune} and \textbf{LoRA}). For example, on the single-label \dataset{MTMD} dataset, DINOv3-L achieves only 78.03\% accuracy when frozen, which surges to 91.67\% with full fine-tuning---an absolute gain of over 13 points. Similarly, on the multi-label \dataset{OralXrays-9}, the mAP for DINOv3-S jumps from 78.99\% (Frozen) to 95.08\% (Finetune). This unequivocally demonstrates that while frozen DINOv3 features provide a strong foundation, task-specific adaptation is essential to unlock their full potential for precise dental image classification, regardless of the imaging modality.

The \textbf{LoRA} strategy emerges as a competitive and efficient alternative to full fine-tuning. On panoramic datasets, LoRA matches or often surpasses the performance of Finetune while using orders of magnitude fewer trainable parameters (TrP). For DINOv3-L on \dataset{OralXrays-9}, LoRA (3M TrP) achieves an mAP of 95.83\%, outperforming Finetune (303M TrP, 95.47\% mAP). On the more complex intraoral dataset \dataset{AlphaDent}, LoRA consistently delivers superior or competitive results; for DINOv3-S, LoRA achieves the highest mAP (68.19\%) and F1-score (59.15\%), significantly outperforming Finetune. This suggests that LoRA may provide more than parameter savings alone, and may also offer a favorable regularizing effect, potentially leading to better generalization, especially on smaller or more challenging datasets like \dataset{AlphaDent}.

In addition, the macro Precision and Recall scores reveal an instructive trade-off modulated by the tuning strategy. Full fine-tuning often pushes for higher \textbf{Recall} at the potential cost of \textbf{Precision}. For instance, with DINOv3-L on \dataset{OralXrays-9}, Finetune yields a Recall of 97.48\% but a Precision of 77.32\%, whereas LoRA offers a more balanced profile (Precision: 78.21\%, Recall: 97.36\%). Conversely, on \dataset{ADCD}, LoRA for DINOv3-B achieves a superior balance, leading to the highest F1-score (65.36\%). This suggests that LoRA's constrained updates may help maintain a better precision-recall equilibrium, which is critical for clinical applications where both false positives and false negatives carry costs.

Therefore, for panoramic radiograph classification, LoRA provides strong performance on both single-label and multi-label tasks with high parameter efficiency, making it an attractive option for developing and deploying multiple specialized models. For intraoral photograph classification, LoRA also remains a competitive choice, particularly for complex multi-label tasks. Full fine-tuning can be considered if the sole objective is to maximize mAP.
The Frozen strategy is not recommended for final deployment given its significant performance deficit, though it remains a useful baseline for rapid prototyping.

\begin{TabTwo}[t]

\caption{Classification results with Frozen / Fine-tuning / LoRA on three panoramic datasets.
}

\label{tab:cls_3datasets_frozen_finetune_lora_n8}

\begin{adjustbox}{max width=\linewidth}
\begin{tabular}{l l l cccc cccc cccc}
\toprule
\textbf{Backbone} & \textbf{Tuning} & \textbf{TrP} &
\multicolumn{4}{c}{\textbf{MTMD (Single-label)}} &
\multicolumn{4}{c}{\textbf{OralXrays-9 (Multi-label)}} &
\multicolumn{4}{c}{\textbf{ADCD (Multi-label)}} \\
\cmidrule(lr){4-7}\cmidrule(lr){8-11}\cmidrule(lr){12-15}
& & &
\textbf{Acc}  & \textbf{Prec.}  & \textbf{Rec.}  & \textbf{F1}  &
\textbf{mAP}  & \textbf{Prec.}  & \textbf{Rec.}  & \textbf{F1}  &
\textbf{mAP}  & \textbf{Prec.}  & \textbf{Rec.}  & \textbf{F1}  \\
\midrule

\multirow{3}{*}{\bk{DINOv3-S}{dinov3}}
& Frozen   & 5K   & 78.03 & 37.89 & 42.40 & 39.98  & 78.99 & 56.16 & 96.08 & 68.95  & 64.84 & 41.52 & 78.50 & 52.45 \\
& Finetune & 21M  & 88.26 & 80.37 & 77.68 & 77.55  & 95.08 & \best{78.86} & 96.44 & \best{85.95}  & 73.06 & 55.05 & 79.83 & 63.69 \\
& LoRA     & 595K & 88.64 & 77.07 & 82.86 & 78.63  & 95.47 & 77.73 & 97.24 & 85.45  & 70.90 & 54.69 & 78.90 & 63.29 \\
\midrule

\multirow{3}{*}{\bk{DINOv3-B}{dinov3}}
& Frozen   & 10K  & 76.89 & 42.14 & 42.75 & 42.19  & 79.79 & 55.93 & 95.77 & 69.18  & 64.86 & 42.82 & 79.00 & 54.41 \\
& Finetune & 85M  & 89.02 & 88.05 & 69.62 & 73.05  & 95.50 & 77.59 & 96.62 & 85.12  & 70.29 & 54.85 & 84.35 & 64.43 \\
& LoRA     & 1M   & 90.15 & 79.59 & 79.77 & 79.68  & 95.56 & 76.22 & \best{97.68} & 84.51  & \best{74.83} & 56.77 & \best{84.69} & 65.36 \\
\midrule

\multirow{3}{*}{\bk{DINOv3-L}{dinov3}}
& Frozen   & 14K  & 78.03 & 37.62 & 43.85 & 40.43  & 80.22 & 57.50 & 96.12 & 70.13  & 62.83 & 44.61 & 76.76 & 55.09 \\
& Finetune & 303M & \best{91.67} & \best{88.54} & \best{84.30} & \best{85.44}  & 95.47 & 77.32 & 97.48 & 85.03  & 72.54 & \best{57.07} & 74.86 & 64.16 \\
& LoRA     & 3M   & 90.53 & 78.29 & 83.98 & 80.73  & \best{95.83} & 78.21 & 97.36 & 85.82  & 72.05 & 56.54 & 84.67 & \best{65.63} \\
\bottomrule
\end{tabular}%

\end{adjustbox}
\end{TabTwo}

\begin{table}[t]
\centering
\scriptsize
\setlength{\tabcolsep}{2.4pt} 
\renewcommand{\arraystretch}{1.05}

\caption{Multi-label classification results with Frozen / Fine-tuning / LoRA  on two intraoral datasets.
}

\label{tab:ml_cls_2datasets_frozen_finetune_lora_n8}

\resizebox{\columnwidth}{!}{%
	\begin{tabular}{ll cccc cccc}
\toprule
\textbf{Backbone} & \textbf{Tuning} &
\multicolumn{4}{c}{\textbf{DentalAI (Multi-label)}} &
\multicolumn{4}{c}{\textbf{AlphaDent (Multi-label)}} \\
\cmidrule(lr){3-6}\cmidrule(lr){7-10}
& &
\textbf{mAP}  & \textbf{Prec.}  & \textbf{Rec.}  & \textbf{F1}  &
\textbf{mAP}  & \textbf{Prec.}  & \textbf{Rec.}  & \textbf{F1}  \\
\midrule

\multirow{3}{*}{\bk{DINOv3-S}{dinov3}}
& Frozen   & 76.02 & 61.53 & 84.49 & 70.14 & 58.65 & 49.46 & 71.46 & 52.61 \\
& Finetune & 82.52 & 72.16 & 83.10 & 76.99 & 63.18 & 45.48 & 70.54 & 54.21 \\
& LoRA     & 84.34 & 76.39 & 82.30 & 79.16 & 68.19 & 51.42 & 70.82 & 59.15 \\
\midrule

\multirow{3}{*}{\bk{DINOv3-B}{dinov3}}
& Frozen   & 74.16 & 60.99 & 83.32 & 69.43 & 58.98 & 43.05 & 70.44 & 51.98 \\
& Finetune & 84.49 & 71.15 & 86.78 & 77.68 & 69.65 & 54.51 & 65.59 & \best{59.18} \\
& LoRA     & 85.04 & 74.86 & 85.77 & 79.70 & \best{69.73} & 49.63 & 71.98 & 58.21 \\
\midrule

\multirow{3}{*}{\bk{DINOv3-L}{dinov3}}
& Frozen   & 74.95 & 60.60 & 82.21 & 68.71 & 53.24 & 39.94 & \best{72.01} & 50.10 \\
& Finetune & 86.81 & \best{79.32} & 87.38 & \best{83.04} & 66.16 & 49.21 & 70.23 & 56.73 \\
& LoRA     & \best{86.91} & 72.64 & \best{89.88} & 79.53 & 67.40 & \best{54.96} & 63.98 & 59.02 \\
\bottomrule
\end{tabular}%

}
\end{table}

\section{Limitations of this benchmark}
While this report presents a comprehensive benchmark across diverse dental analytic tasks and data modalities, it has several limitations.
First, some classification settings are derived by aggregating detection or segmentation annotations, and therefore may not fully match clinically curated image-level labels.
Second, although we standardize training budgets and compare backbones under unified protocols, the downstream architectures themselves (e.g., Mask R-CNN and UPerNet) introduce inductive biases that may influence absolute performance, even if the relative trends are still informative.
Third, our benchmark is limited to 2D dental imagery, including radiographs and intraoral photographs; whether the same conclusions transfer to 3D CBCT, multi-view studies, or longitudinal dental records remains an open question.
Finally, dental datasets often include site- and device-specific shifts as well as annotation variability, and understanding how these factors interact with pretraining and adaptation warrants further study.


\section{Conclusion}\label{sec:conclusion}
In this work, we introduce \model{DinoDental}, a comprehensive benchmark designed to evaluate the transferability of the \model{DINOv3} foundation model across the diverse tasks and modalities within dental imaging. Through systematic evaluation spanning classification, object detection, and segmentation, we draw three principal conclusions regarding the adoption of such models in dentistry.

First, DINOv3 proves to be a robust, general-purpose visual encoder, exhibiting particular strength in texture-rich intraoral image analysis and complex segmentation tasks. However, its advantage is not absolute; for certain tasks, such as the detection of well-defined anatomical structures in panoramic radiographs, conventional CNNs can remain competitive. This indicates that backbone selection should be carefully aligned with the specific imaging modality and clinical task.
Second, model performance is strongly influenced by input resolution, with gains typically plateauing at approximately 1024×1024 pixels. Beyond this point, the benefit of increased spatial detail is counterbalanced by the amplification of inherent imaging noise, particularly in X-ray modalities, yielding diminishing returns and advising against indiscriminate image upscaling.
Third, and most critically, LoRA emerges as the most practical adaptation strategy under our benchmark setting, offering the strongest accuracy--efficiency trade-off and often matching or surpassing full fine-tuning. It effectively bridges the domain gap between natural-image pre-training and medical imaging while concurrently regularizing the model to mitigate overfitting. This positions LoRA as a highly efficient and scalable method for harnessing large-scale pre-trained models.



{\small
\bibliographystyle{ieee_fullname}
\bibliography{egbib}
}

\end{document}